\documentclass[10pt,journal,compsoc]{IEEEtran}

\usepackage{ragged2e}
\usepackage{graphicx}
\usepackage{amsmath}
\usepackage{amssymb}
\usepackage{booktabs}
\usepackage[hidelinks]{hyperref}
\usepackage{url}
\usepackage{multirow}
\usepackage{array}
\usepackage{tabularx}
\usepackage{float}
\usepackage{capt-of}
\usepackage{longtable}

\ifCLASSOPTIONcompsoc
  \usepackage[nocompress]{cite}
\else
  \usepackage{cite}
\fi

\ifCLASSINFOpdf
\else
\fi

\newcommand{\indicator}[1]{\mathbf{1}\!\left\{#1\right\}}

\begin{document}
\title{Code2Worlds-2: Physically Plausible Agent Harness for Embodied World Simulation}

\author{Haoyu Zhang$^*$,
        Siming Li$^*$,
        Zeyu Zhang$^{*\dag}$,
        Yi Zhang,
        Tat-Seng Chua,
        Hao Tang$^\ddag$
\IEEEcompsocitemizethanks{
        \IEEEcompsocthanksitem $^*$Equal contribution. $^\dag$Project lead. \protect
        \IEEEcompsocthanksitem $^\ddag$Corresponding author, E-mail: bjdxtanghao@gmail.com. \protect
        \IEEEcompsocthanksitem Haoyu Zhang, Siming Li, Zeyu Zhang, Yi Zhang, and Hao Tang are with the School of Computer Science, Peking University, Beijing 100871, China. \protect
        \IEEEcompsocthanksitem Tat-Seng Chua is with the School of Computing,
National University of Singapore, Singapore.
    }%
}

\markboth{Submitted to IEEE Transactions on Pattern Analysis and Machine Intelligence}%
{Zhang \MakeLowercase{\textit{et al.}}: Code2Worlds-2: Physically Plausible Agent Harness for Embodied World Simulation}

\IEEEtitleabstractindextext{%
\begin{abstract}\justifying
Recent coding large language models (LLMs) can synthesize dynamic 4D worlds from natural language, but the resulting content remains a \emph{visual} artifact. It lacks physical semantics such as the collision geometry, mass, friction, inertia, and joint structure, required to enable it to simulate the physical world, and its dynamics are scripted to \emph{look} plausible rather than designed to \emph{execute}. What is needed is an \emph{executable} world: one that a physics engine can load and step, resolving contact physically.
 We present \textbf{Code2Worlds-2}, an extension of Code2Worlds that turns 4D world generation into an \emph{embodied world simulation}. We keep the original dual-stream generation pipeline intact and append two modules. First, a \textbf{SimReady Scene Compiler} that turns a generated visual scene into a simulator-compatible representation: it assigns collision meshes via convex decomposition, estimates physical attributes with a vision--language model, infers a lightweight support graph, instantiates the engine's contact, friction, and dynamics constraints, and applies a gravity settling pass, before exporting the scene as MuJoCo MJCF. Second, a \textbf{Simulator-in-the-Loop 4D Refiner} that implements an automated \emph{agent harness}---a compile, simulate, critique and refine loop---that converts visual dynamics into executable rigid and articulated dynamics and evolves the \emph{physical} settings rather than only the code, replacing the purely visual motion critic with a simulator critic that inspects penetration depth, collision pairs, stability, contact forces, joint limits, and robot reachability. We further propose \textbf{SimCode4D}, a state-based benchmark with five task templates for evaluating embodied simulation. Experiments on SimCode4D and across natural, indoor, industrial, urban, and service scenes demonstrate that Code2Worlds-2 produces scenes that are physically plausible, simulator-compatible, and interactable by legged and manipulator robots.
Code: \url{https://github.com/AIGeeksGroup/Code2Worlds-2}.
Website: \url{https://aigeeksgroup.github.io/Code2Worlds-2}.
\end{abstract}

\begin{IEEEkeywords}
4D world generation, embodied world simulation, coding LLM, physics simulation, robot manipulation, MuJoCo.
\end{IEEEkeywords}}

  \maketitle

\IEEEdisplaynontitleabstractindextext

\IEEEpeerreviewmaketitle

\IEEEraisesectionheading{\section{Introduction}\label{sec:introduction}}

\IEEEPARstart{S}{patial} intelligence requires more than visually convincing pictures; it demands world models that obey the laws of physics. Coding large language models (LLMs) have recently been shown to generate static 3D scenes and, more ambitiously, dynamic 4D worlds by writing programmatic scene code. Code2Worlds~\cite{code2worlds} advanced this line by proposing a dual-stream architecture that disentangles retrieval-augmented object generation from hierarchical environmental orchestration, and by narrowing the semantic--physical gap with a Post-Process Agent that scripts dynamics and a VLM-Motion Critic that self-reflects to refine the generated code. The result is a 4D world that is \emph{visually} dynamic and \emph{semantically} physical.

\begin{figure}[t]
\centering
\includegraphics[width=\linewidth]{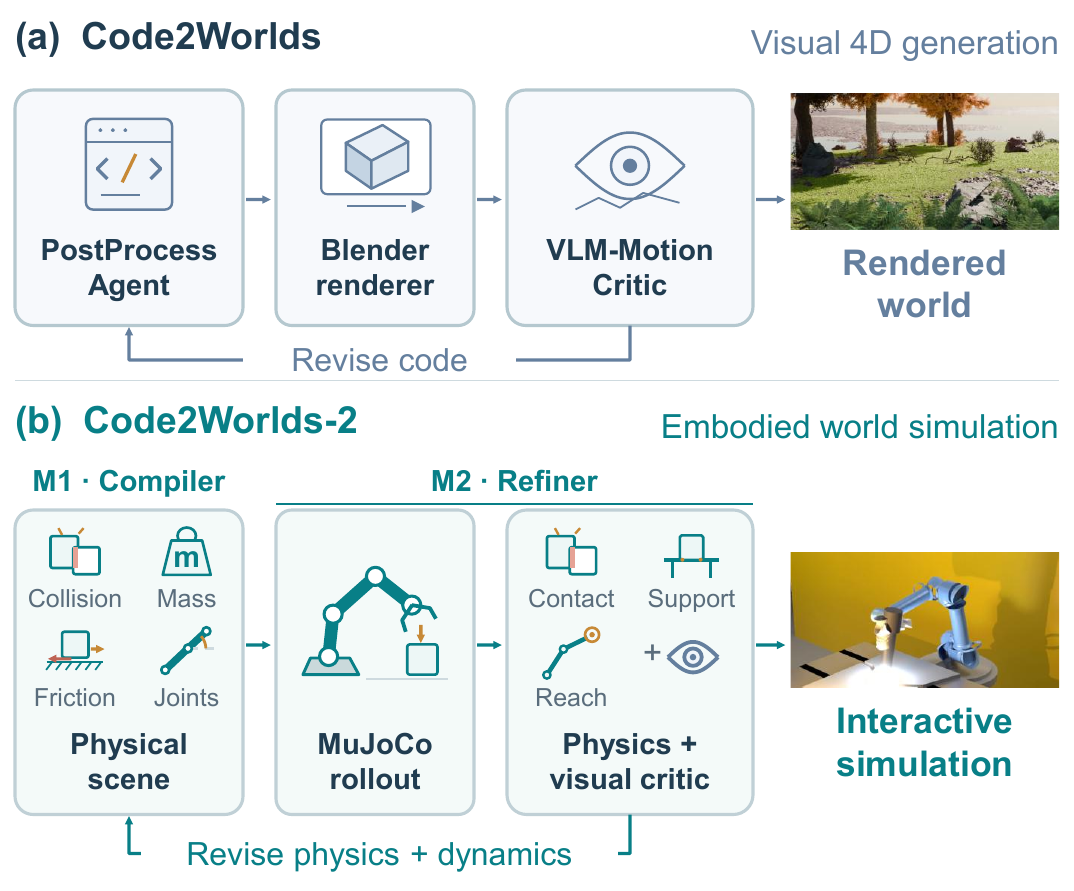}
\caption{The feedback loops of the two systems. (a)~Code2Worlds~\cite{code2worlds} corrects a scene by rewriting its code from rendered images---a \emph{visual} loop, which cannot guarantee physical validity. (b)~Code2Worlds-2 instead closes the loop through the simulator: Module~1 compiles a SimReady scene, Module~2 critiques the rollout, and simulator feedback guides the revisions to the \emph{physical settings} and dynamics.}
\label{fig:comparison}
\end{figure}

Yet for an embodied agent---a quadruped that must walk across a rocky field, a manipulator that must open a drawer---a visually plausible world is not enough. Two gaps separate Code2Worlds from a usable embodied simulator. First, a \textbf{semantic--physical execution gap}: generated meshes carry no collision geometry, mass, friction, inertia, or joint structure, so a physics engine such as MuJoCo~\cite{todorov2012mujoco} or Isaac Sim~\cite{isaacsim} accepts them only as raw visual hulls, where contact, support, and manipulation carry no physical meaning. Second, a \textbf{simulation--fidelity gap}: the dynamics are scripted to \emph{appear} physically reasonable but are never executed, so an agent cannot collide with, push, grasp, or otherwise interact with them. Fig.~\ref{fig:comparison} contrasts the two loops: Code2Worlds rewrites the entire scene code, whereas Code2Worlds-2 revises the physical settings themselves.

\begin{figure*}[t]
\centering
\includegraphics[width=\textwidth]{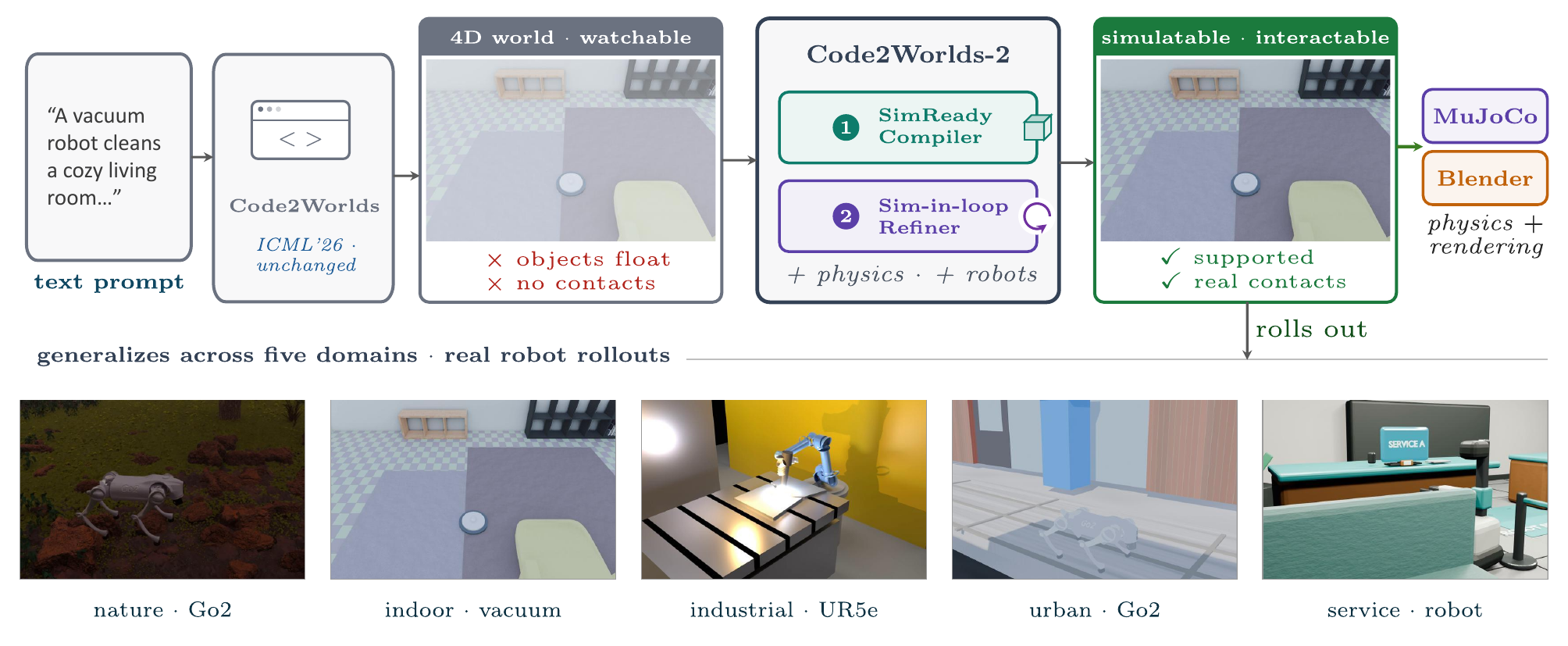}
\caption{Overview of Code2Worlds-2. Two modules are appended to the original Code2Worlds pipeline to produce a simulator-compatible, interactable 4D world, and the illustrated scenes span five domains.}
\label{fig:overview}
\end{figure*}

This paper extends Code2Worlds into \textbf{Code2Worlds-2}, an \emph{embodied world simulation} framework. We keep the original pipeline unchanged and append two modules that bridge these gaps (Fig.~\ref{fig:overview}). The first, a \textbf{SimReady Scene Compiler} (Module~1), wraps each visual scene with a physical compilation layer: it decomposes concave meshes into convex pieces, estimates physical attributes with a vision--language model and geometry, infers a support graph, and settles objects under gravity before exporting MJCF. The second, a \textbf{Simulator-in-the-Loop 4D Refiner} (Module~2), translates visual dynamics into executable dynamics and replaces the purely visual motion critic with a simulator critic that reasons over physical signals rather than pixels alone. We additionally propose \textbf{SimCode4D}, a state-based benchmark with five task templates, and validate the framework across natural, indoor, industrial, urban, and service scenes.

\textbf{Extension over the conference version.} This paper extends our ICML 2026 conference paper, Code2Worlds~\cite{code2worlds}. The original generation streams are retained; the journal version adds (i) physical scene compilation and MuJoCo export (Sec.~\ref{sec:module1}), (ii) simulator-in-the-loop dynamics refinement and physical validation (Sec.~\ref{sec:module2}), and (iii) the state-based SimCode4D protocol and embodied-interaction evaluation (Secs.~\ref{sec:benchmark} and~\ref{sec:experiments}).

In summary, this paper makes the following contributions:
\begin{itemize}
    \item We propose a \textbf{SimReady Scene Compiler} that converts coding-LLM-generated visual scenes into simulator-compatible representations with collision geometry, physical attributes, a support graph, physical constraints, and a settling pass, while emitting paired \emph{default} and \emph{enriched} scene exports whose comparison is our primary measure of physical plausibility.
    \item We propose an \textbf{agent harness}---an automated compile--simulate--critique--refine loop that evolves the \emph{physical} settings of a generated world---realized by a \textbf{Simulator-in-the-Loop 4D Refiner} that compiles visual dynamics into executable rigid and articulated dynamics and replaces visual-only critique with simulator-signal critique (penetration, contact, stability, joint limits, reachability).
    \item We introduce \textbf{SimCode4D}, a state-based benchmark with five task templates for embodied world simulation, and demonstrate physical plausibility, simulator compatibility, and robot interactability across five scene categories.
\end{itemize}

\section{Related Work}\label{sec:related}

\subsection{Coding LLMs for Scene Generation}
Recent work shows that coding LLMs can generate 3D scenes by emitting programmatic code rather than raw assets, benefiting from the compositionality and editability of code. SceneCraft~\cite{hu2024scenecraft} drives an LLM agent that writes Blender code and refines it from rendered feedback; LayoutGPT~\cite{feng2023layoutgpt} casts scene composition as language-conditioned layout planning; 3D-GPT~\cite{sun2023dgpt} decomposes procedural modeling into language-driven tasks; and BlenderLLM~\cite{du2024blenderllm} trains a model specifically for computer-aided design code. Related code-generative efforts target individual assets rather than whole scenes, including LL3M~\cite{lu2025ll3m} and MeshCoder~\cite{dai2025meshcoder}. SceneCode~\cite{wang2026scenecode} compiles articulated indoor scenes into executable world programs, starting from existing scenes rather than a coding-LLM-generated 4D world. Code2Worlds~\cite{code2worlds} extends this line to 4D by proposing a dual-stream architecture and a physics-aware closed loop. Our work builds directly on Code2Worlds and targets the step its design does not address: making the generated world \emph{simulator-compatible} and its dynamics \emph{executable}.

\subsection{Physics-Aware Scene Construction}
Several concurrent systems make generated or reconstructed scenes physically plausible, and we build on the physical-property conventions they established.

\noindent\textbf{SceneSmith}~\cite{scenesmith} generates indoor scenes hierarchically through a Designer--Critic--Orchestrator loop. It combines generated assets, articulated-object libraries, and thin coverings, estimates physical properties with a vision--language model (VLM), and builds V-HACD collision geometry. Collision and reachability checks guide generation, followed by collision-free projection and gravity settling in Drake~\cite{drake}. It is the closest prior work in spirit, but generates scenes from scratch rather than compiling an existing coding-LLM world.

\noindent\textbf{InternScenes}~\cite{internscenes} assembles simulatable indoor scenes from real scans, procedural Infinigen~\cite{infinigen} outputs, and cleaned designer scenes. It optimizes large-furniture bounding boxes to reduce overlap and ground misalignment, then settles small objects in SAPIEN. It emphasizes realistic layouts and collision elimination, with lighter physical-property modeling than SceneSmith.

\noindent\textbf{SimRecon}~\cite{simrecon} reconstructs object-level, composable, simulator-ready scenes from real RGB video. It first performs semantic 3D reconstruction (2DGS plus SceneSplat segmentation), then completes each object with an active-viewpoint optimization (AVO), and finally assembles the scene through a Scene Graph Synthesizer (SGS) that infers $(child, relation, parent)$ triples---primarily \texttt{supported\_by} and \texttt{attached\_to}---via VLM over instance-ID renders, followed by hierarchical physical assembly: support relations settle objects under gravity, attachment relations apply fixed constraints. SimRecon builds a support graph similar to ours, but it operates on reconstruction rather than generation.

These systems each address one segment of ``physical scene construction,'' but none takes a coding-LLM-generated \emph{dynamic 4D world} as input and compiles it end-to-end into a representation that a physics simulator can load and a robot can act on. Code2Worlds-2 occupies exactly this position: it inherits Code2Worlds's dual-stream generation and Post-Process Agent, and adds a compiler (Module~1) and a simulator-in-the-loop refiner (Module~2) that make the generated world physically grounded and robot-interactable. Moreover, our settling pass produces simulator contact data that can ground the support graph in measured contact, rather than in VLM or geometric heuristics alone.

\noindent\textbf{Other physically grounded generators.} Holodeck~\cite{yang2024holodeck} generates embodied 3D environments from language descriptions, while PhyScene~\cite{yang2024physcene} synthesizes physically interactable scene layouts from a given floor plan and room type by guiding a conditional diffusion model with collision, room-boundary, and reachability constraints, and RoboGen~\cite{wang2024robogen} generates tasks together with their simulation environments. PAT3D~\cite{lin2026pat3d} augments text-to-3D generation with physical reasoning, while URDFormer~\cite{chen2024urdformer} predicts articulated structure and URDF descriptions from images. Common asset and scene sources for these systems include Objaverse~\cite{deitke2023objaverse}, Infinigen Indoors~\cite{raistrick2024infinigenindoors}, and CityX~\cite{zhang2024cityx}.

\subsection{Simulation Platforms and Robot Benchmarks}
Our compiler targets an explicit engine representation supported by several platforms: MuJoCo~\cite{todorov2012mujoco}, Isaac Gym~\cite{makoviychuk2021isaacgym}, Isaac Lab~\cite{mittal2025isaaclab}, Brax~\cite{freeman2021brax}, and SAPIEN~\cite{xiang2020sapien}. Manipulation benchmarks such as RLBench~\cite{james2020rlbench}, ManiSkill2~\cite{gu2023maniskill2}, robosuite~\cite{zhu2020robosuite}, Meta-World~\cite{yu2019metaworld}, CALVIN~\cite{mees2022calvin}, and BEHAVIOR-1K~\cite{li2024behavior1k}, and interactive suites such as AI2-THOR~\cite{kolve2017ai2thor}, Habitat~\cite{savva2019habitat}, RoboTHOR~\cite{deitke2020robothor}, ProcTHOR~\cite{deitke2022procthor}, Gibson~\cite{xia2018gibson}, and HM3D~\cite{ramakrishnan2021hm3d}, all take simulator-ready assets as input. SimCode4D instead asks whether an LLM-generated world can \emph{become} one.

\subsection{Generative World Models}
A separate line learns interactive dynamics directly from video, including Genie~\cite{bruce2024genie}, UniSim~\cite{yang2024unisim}, GAIA-1~\cite{hu2023gaia1}, WorldDreamer~\cite{wang2024worlddreamer}, and Ctrl-World~\cite{guo2025ctrlworld}. Such models can render plausible futures, but they do not expose contact, friction, or joint state, so contact-level physical validation cannot be performed on them. Code2Worlds-2 instead commits to an explicit, engine-agnostic representation that a simulator can load.

\subsection{Physical Property Estimation and 3D Scene Graphs}
Estimating physical attributes from visual input has been approached with language-embedded features~\cite{zhai2024physical}, single-image reasoning~\cite{le2026siphy,lee2026physically}, and multi-view material decomposition of generated meshes~\cite{comi2026gen2physics}; we adopt this idea as one stage of the compiler rather than as an end in itself. A related literature represents scenes as graphs over objects and relations~\cite{armeni2019scene,wald2020learning,wu2021scenegraphfusion,koch2024open3dsg,gu2024conceptgraphs}; our support graph is a deliberately lightweight, support- and contact-focused instance of that idea.

\section{Method}\label{sec:method}

\begin{figure*}[t]
\centering
\includegraphics[width=\textwidth]{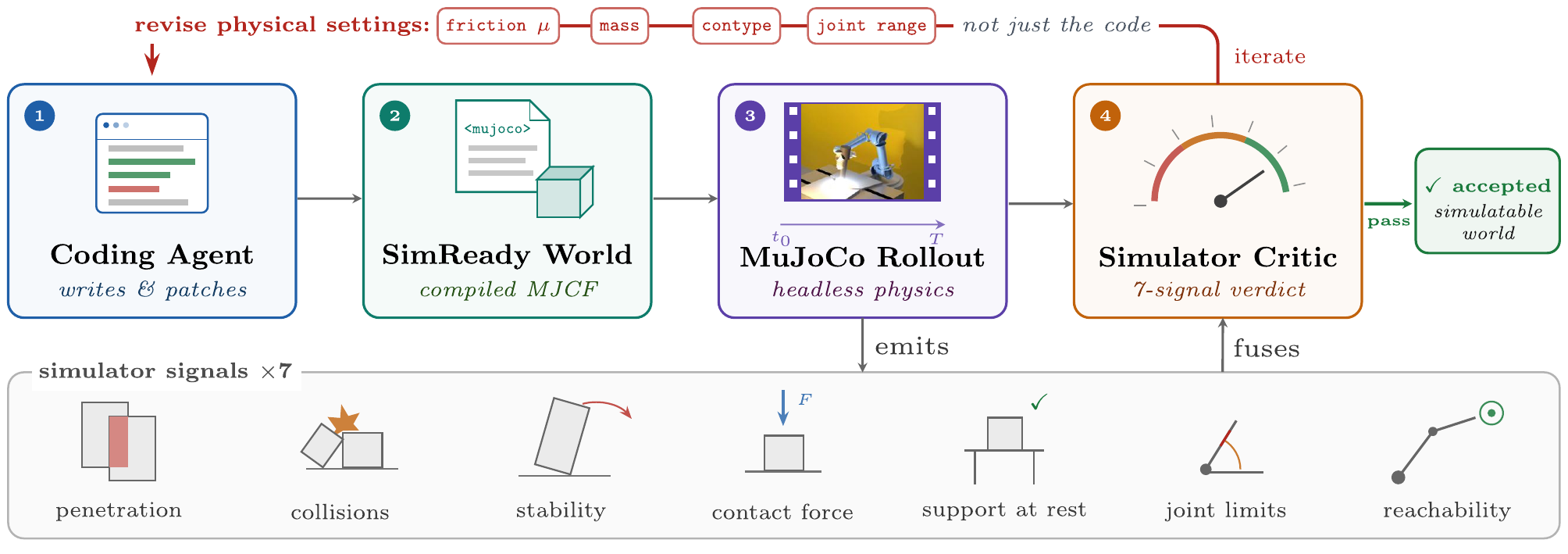}
\caption{The Code2Worlds-2 agent harness. The generation agent closes a loop through the simulator: it emits code and a dynamics program, which are compiled, executed, and measured by physical signals; the fused verdict then revises the \emph{physical} settings until validation passes.}
\label{fig:harness}
\end{figure*}

The central abstraction of Code2Worlds-2 is an \emph{Embodied World Graph}, an engine-agnostic intermediate representation that bridges the visual world emitted by Code2Worlds and a target physics engine (MuJoCo in this work). Each \emph{object node} carries the visual mesh, collision mesh, mass, friction, inertia, center of mass, scale, pose, support/contact relation, affordance, and joint type; each \emph{dynamic event} carries force, velocity, gravity response, contact events, object state transitions, temporal triggers, and interaction constraints. The compiler produces this graph once, then serializes it to a target simulator (here MuJoCo MJCF), so that the same world can in principle be exported to Isaac Sim/USD or other engines without regenerating the scene.

\textbf{Agent harness.} The other key component of Code2Worlds-2 is a \emph{harness}: the automated control loop that lets a coding agent produce a physically executable world without human intervention. In Code2Worlds, the Post-Process Agent refines \emph{code} against a \emph{visual} critic; in Code2Worlds-2, the harness closes the loop through the simulator itself. The agent generates scene code and a dynamics program, Module~1 compiles them into an Embodied World Graph and instantiates the engine's constraints, Module~2 executes the rollout in MuJoCo, the simulator critic measures physical signals (penetration, contact, stability, forces, joint limits, reachability), and the fused verdict is fed back to the agent to revise the \emph{physical settings}---not merely the script. The loop iterates until the scene passes both physical and visual validation (Fig.~\ref{fig:harness}).

We retain the Code2Worlds generation pipeline (dual-stream object/scene generation and the Post-Process Agent), extended so that the agent emits a dynamics program and acts on simulator feedback, and append two modules (Fig.~\ref{fig:overview}): Module~1 (M1) compiles visual scenes into simulator-compatible representations; Module~2 (M2) converts and refines their dynamics using simulator feedback.

\subsection{Module 1: SimReady Scene Compiler}\label{sec:module1}

\begin{figure*}[t]
\centering
\includegraphics[width=\textwidth]{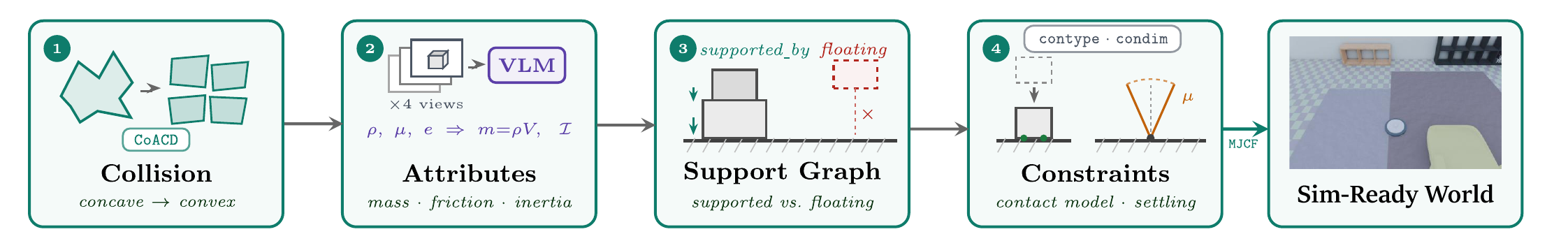}
\caption{The stages of Module~1. Collision geometry is obtained by convex decomposition; density, friction, and restitution are estimated by a vision--language model, while mass and inertia are computed from mesh geometry; a support graph separates supported from floating bodies; and the engine's contact model, together with a gravity settling pass, produces a simulator-compatible scene exported as MJCF.}
\label{fig:module1}
\end{figure*}

Module~1 turns a generated \emph{visual} scene into a \emph{simulation-ready} scene without altering the object or scene streams. Its compilation stages are summarized in Fig.~\ref{fig:module1}.

\subsubsection{Collision Geometry}
A simulator requires collision meshes that are either convex or decomposed into convex pieces; raw generated meshes are concave and non-watertight, and a physics engine will otherwise convex-hull each mesh into a coarse, often wrong solid. We apply CoACD convex decomposition~\cite{coacd} to key interaction objects, with a concavity threshold of $0.05$ and a per-object piece cap (24 in our scenes) to bound solver cost; alternatives include V-HACD~\cite{mamou2016volumetric}, Empart~\cite{vu2025empart}, and VisACD~\cite{fokin2026visacd}. For outdoor terrain, whose non-convex height field would be convex-hulled by MuJoCo into a single solid spanning the whole scene, we slice the terrain into an $N\!\times\!N$ grid and reconstruct, in each cell, a closed convex prism from a ray-cast height field (top surface plus four side walls plus a flat base). Each cell is one geom, so MuJoCo's convex hull follows the local surface, with error bounded by the cell's vertical variation.

\subsubsection{Physical Attribute Estimation}
For every key object we render four views (iso, front, side, top) and send them with the object's longest dimension and source name to a vision--language model, which predicts a material category (e.g.\ granite, steel, plastic), bulk density $\rho$, friction $\mu$, and restitution $e$. We then compute mass from the mesh volume,
\begin{equation}
m = \rho\, V,
\label{eq:mass}
\end{equation}
and the inertia tensor $I$ and center of mass from the mesh moments, following the mass--inertia convention of prior physics-aware systems~\cite{scenesmith}; related visual property estimation includes GaussianProperty~\cite{xu2025gaussianproperty} and PUGS~\cite{shuai2025pugs}. This yields a complete physical-attribute tuple (visual mesh, collision mesh, mass, friction, restitution, inertia, center of mass) without re-authoring any asset.

\subsubsection{Support Graph}
We infer a lightweight support graph $G=(O,R)$ over objects. $R$ contains the structural relations $\{$\texttt{supported\_by}, \texttt{on\_surface}, \texttt{attached\_to}, \texttt{inside}, \texttt{hinged\_to}$\}$ and the settling triggers $\{$\texttt{floating}, \texttt{embedded}$\}$. Relations are inferred from world-space bounding boxes. Support is determined by positive xy-footprint overlap and a vertical gap between an object's bottom and a candidate support surface (floor, terrain, or an object's top): a gap within tolerance yields \texttt{supported\_by}/\texttt{on\_surface}, a large positive gap \texttt{floating}, and a large negative gap \texttt{embedded}. Attachment is detected by face-to-face proximity to a wall with sufficient vertical overlap; containment by full bounding-box enclosure of one object inside another; articulation by naming cues (door, drawer, panel) plus a nearby structural neighbor. Because procedural indoor scenes often export only decorative floor tiles rather than the true room floor, we add a global-ground fallback that recovers the floor height from the lowest credible object bottom, preventing spurious \texttt{floating} labels. The graph drives the settling pass below and is later refined by simulator contact data.

\subsubsection{Physics Settling}
We load each object into MuJoCo slightly above its support surface (5\,cm) and simulate under gravity until it comes to rest, recording the settled pose; an object is accepted as settled when its linear and angular speeds fall below small thresholds and it has at least one contact. Objects labeled \texttt{floating} drop onto their support, objects labeled \texttt{embedded} are pushed out, and objects that still fail to settle are re-placed. The settling pass both corrects the initial scene and serves as a source of contact data that we later exploit to verify the support graph.

\subsubsection{Physical Constraints}
\begin{figure}[t]
\centering
\includegraphics[width=\linewidth]{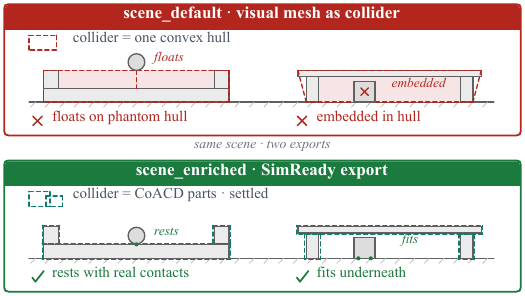}
\caption{Schematic comparison of default (top) and SimReady (bottom) scene exports. A single convex hull fills concave spaces, creating phantom support and embedding the box. CoACD colliders preserve these spaces, allowing the sphere to rest on the seat after settling and the box to fit beneath the table. Dashed outlines denote collision geometry; green dots mark contact points.}
\label{fig:simready-export-comparison}
\end{figure}

Beyond geometry and attributes, a simulator-compatible scene must encode the constraints that make contact and motion physically meaningful. We instantiate three families of constraints in the target engine (MuJoCo). (i) \emph{Contact constraints}: every collider is assigned a \texttt{contype}/\texttt{conaffinity} bitmask so that only intended pairs (e.g., object--floor, robot--object) generate contacts, and \texttt{condim}=3 for normal contact and tangential friction, or \texttt{condim}=4 when torsional friction about the contact normal is also required. (ii) \emph{Friction constraints}: each contact geom carries friction coefficients that, together with \texttt{condim} and the solver's friction cone, bound the enabled tangential forces and frictional torques. (iii) \emph{Dynamics constraints}: unattached rigid bodies use free joints, while hinge and slide joints carry their own axis, range limits, and---for articulated objects---stiffness and damping, so a drawer opens only along its rail and a robot joint respects its limit. The integrator, solver, and gravity are configured once per scene, and the settling pass above runs under exactly these constraints, so the exported MJCF is the same representation that is validated. These constraints complement the collision geometry illustrated in Fig.~\ref{fig:simready-export-comparison}.

Module~1 emits two scene variants: \texttt{scene\_default}, which uses visual meshes as colliders exactly as MuJoCo would by default, and \texttt{scene\_enriched}, which applies the full collision plan (convex decomposition, terrain grid, physical attributes). The gap between the two is our primary measure of physical plausibility (Sec.~\ref{sec:experiments}).

\subsection{Module 2: Simulator-in-the-Loop 4D Refiner}\label{sec:module2}

Module~2 closes the loop: it converts the Post-Process Agent's visual dynamics into simulator-executable dynamics and feeds measured physics back so the \emph{physical settings}---not only the code---are revised. Motion is partitioned into two layers: (i) \emph{simulator state}---rigid-body and articulated motion---executed by MuJoCo; and (ii) \emph{visual effects}---rain, fog, fire, relighting---which remain Blender-level dynamics. The Post-Process Agent therefore emits a \texttt{dynamics\_program.json} with three block types---rigid-body (pose, velocity, attributes), joint (definition and trajectory), and visual-effect (free-form descriptor for Blender).

For each dynamic body we resolve mass, friction, and inertia from the physical attributes of Module~1 (or from a material lookup for unambiguously named objects), snap it above the ground or floor surface by ray-casting, and emit a MuJoCo body with a free joint; articulated objects are emitted with a slide or hinge joint driven by a time-parameterized trajectory. A robot (Unitree Go2, UR5e, Clearpath Jackal, or the service robots' differential-drive base) is placed relative to the target according to the action---a kick anchors the foot reach, a walk provides a runway, and a watch observes from a fixed offset---and controlled by PD tracking to kick, walk, watch, or weld. Per frame we log the object pose, speeds, contact count, and peak contact force for the critic below.

\subsubsection{Simulator-in-the-Loop Critic}
We replace the original visual-only VLM-Motion Critic with a two-part critic. The \emph{physics critic} computes seven simulator signals over the rollout log, one per validation objective: penetration depth (maximum sink below the initial support), collision pairs (time-averaged and peak contact count), post-strike stability (mean settle-window speed, which must decay to near zero), contact-force peak (peak normal force during the strike), support relation at rest (final pose on the ground and at rest), joint-limit/uprightness (limit violations or base tip-over), and interaction success (whether the action displaced the object). The \emph{visual critic} asks a VLM whether sampled rollout frames look plausible, returning a validity flag and free-text issues. The two are fused into a single verdict: a scene is accepted only if every physics signal passes the checks above and the VLM judges the sampled frames valid. The fused feedback is returned to the Post-Process Agent, which revises the dynamics program \emph{and} the physical settings that the critic flagged, so the refinement evolves the physics rather than the script alone.

\section{SimCode4D Benchmark}
\label{sec:benchmark}

\subsection{Evaluation Target and Frozen Cases}

We introduce \textsc{SimCode4D}, a state-based benchmark that evaluates
executable world packages, rather than rendered videos, for simulator
execution, physical admissibility, and embodied task completion. Five
templates---pick-and-place, push-to-region, place-into-container,
obstacle-avoidance navigation, and dynamic interaction---use typed
initial, event, always, and terminal predicates. The frozen suite
instantiates three task families: navigation, dynamic interaction, and
pick-and-transport, a pick-and-place variant with target-region transfer.

We instantiate the benchmark with a frozen 15-case development suite,
hereafter \textsc{Dev15}. It contains three world--task cases in each
of five domains: indoor, industrial, natural outdoor, service, and
urban. Its seven navigation, two pick-and-transport, and six
dynamic-interaction cases cover cargo retention, welding,
contact-before-motion, water-to-land transition, drawer interaction,
and contact-mediated blocking.

Each case specifies a goal, world identifier, robot and initial state,
required entities, evaluator-owned predicates, contact permissions,
observations, timeout, and terminal tolerances. Predicates read MuJoCo
state, contact, control, and event logs. Four machine-readable
contracts---\texttt{case\_spec.json}, \texttt{case\_gold.json},
\texttt{physics\_profile.json}, and \texttt{task\_confirm.yaml}---are
paired with an independent probe and validator. Twelve cases retain
authoring scenes; all 15 retain compiled M1 worlds and source pointers.

\subsection{Coding Evaluation}

For the fixed-source track, a method receives the same frozen source bundle and
must emit (i) a world package containing MJCF, local assets, physical fields,
collision geometry, articulations, and persistent entity bindings, and (ii) a
typed task-dynamics overlay. The evaluator owns the robot probe, so a submitted
method cannot encode the evaluation policy in its world program.

For a first-attempt world package $w$, the strict build gate is
\begin{equation}
B_w=P_w\wedge S_w\wedge A_w\wedge E_w\wedge L_w\wedge F_w\wedge I_w,
\label{eq:simcode4d-build-gate}
\end{equation}
where $P$ denotes syntax parsing, $S$ strict schema and cross-reference
validity, $A$ closure and consistency of local assets, $E$ coverage of
required entities, relations, physical attributes, collision geometry,
articulations, and observations, $L$ simulator compilation and loading,
$F$ finite smoke execution, and $I$ intervention integrity. The primary
method-level coding metric is
\begin{equation}
\mathrm{WorldBuildPass@1}
=\frac{1}{N_W}\sum_{w=1}^{N_W}\indicator{B_w}.
\label{eq:simcode4d-world-build}
\end{equation}
Beyond \texttt{WorldBuildPass@1}, we report three first-output coding
diagnostics. \texttt{TaskOverlayPass@1} measures the fraction of packages
whose typed overlay passes schema checks, refers to the compiled world,
and declares all registered dynamic bodies. Required-entity binding and initial support-relation micro-F1 are
computed against registered source bindings and human-confirmed
task-object support labels, respectively. For each measure, micro-F1
pools true-positive (TP), false-positive (FP), and false-negative (FN)
counts across eligible first outputs; unannotated background relations
are excluded.
Physical-attribute, collision, and articulation checks are part of
the build gate rather than separate coverage rates. Runtime exceptions
count as build or execution failures, and feedback calls are reported
for repaired outputs. We omit BLEU and CodeBLEU because equivalent
world programs need not share a reference string.

The \textsc{Dev15} preflight checks frozen reference packages, not
generated first outputs. It verifies registered files, contract
bindings, M1 and runtime loading, M1--M2 binding, overlays, contact
groups, observation and actuator names, and probe integrity. Probes
cannot import the simulator or modify state, applied forces, equality
constraints, mass or inertia, friction, or collision masks. These checks
establish evaluator readiness.

\subsection{Physical, Kinematic, and Task Evaluation}

All \textsc{Dev15} runtime models use gravity
$(0,0,-9.81)\,\mathrm{m\,s^{-2}}$, a $1\,\mathrm{ms}$ step, and MuJoCo's
\texttt{implicitfast} integrator. The physical profile retains body mass and
inertia, joint type/range/damping, actuator gear and control limits, collision
masks, semantic contact groups, and sliding, torsional, and rolling friction.
Evaluation separates four complementary components.

\begin{figure*}[t]
    \centering
    \includegraphics[width=0.99\textwidth]{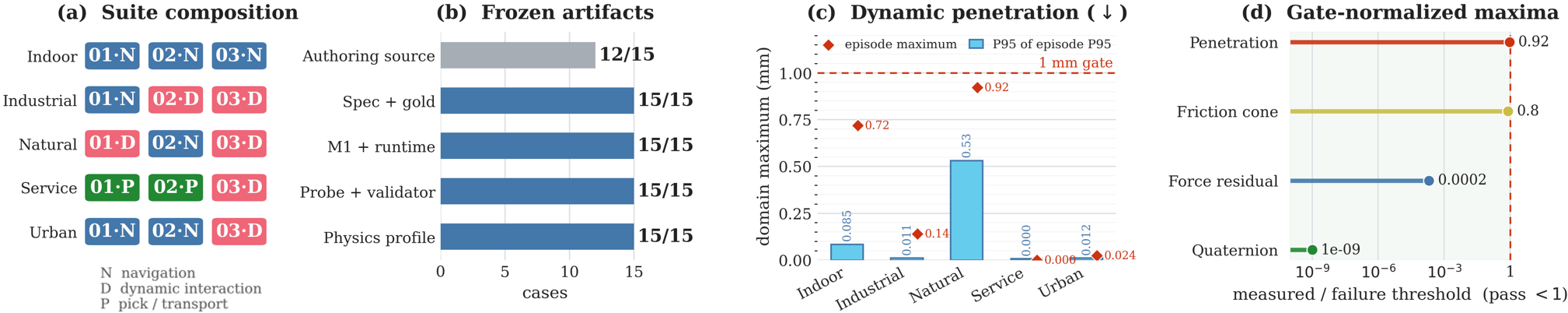}
    \caption{\textsc{Dev15} benchmark composition and execution diagnostics.
    (a) The 15 cases span five domains and three realized task families
    (N: navigation, D: dynamic interaction, P: pick/transport).
    (b) All 15 cases have complete executable contracts; 12 additionally retain
    their authoring scenes.
    (c) For each domain, bars show the 95th percentile of per-episode
    penetration P95 values, whereas diamonds show the maximum penetration
    observed in any episode; both remain below the frozen $1\,\mathrm{mm}$
    gate, and lower is better.
    (d) Maximum penetration, Coulomb utilization, generalized-force residual,
    and quaternion error are normalized by their failure thresholds; lower is
    better, and all values remain below one.}
    \label{fig:simcode4d-dev15-summary}
\end{figure*}

\paragraph*{Mechanics}
We require finite dynamics and a forward/inverse generalized-force residual no
larger than $10^{-5}$. Contact force, kinetic and potential energy, and the
constraint-force residual remain visible diagnostics. The latter is not used
as a substitute for generalized-force balance because its normalization is
solver dependent. Undeclared equality constraints, forbidden contacts, and
action-bound violations are hard failures.

\paragraph*{Kinematics}
We check finite pose and twist, quaternion norm error below $10^{-6}$, joint
range and speed, and finite-difference linear/angular acceleration and jerk.
Case-specific predicates further constrain path corridors, pose, tilt, event
order, goal dwell, and terminal speed.

\paragraph*{Contact and friction}
For every resolved contact, the evaluator records penetration, normal and
tangential force, tangential slip speed, and contact-dimension-aware Coulomb
utilization using the declared sliding, torsional, and rolling coefficients.
A friction-cone failure requires utilization above $1.25$ continuously for at
least $10\,\mathrm{ms}$. Zero-contact episodes remain in the denominator of
friction-contact coverage rather than being imputed.

\paragraph*{Robust task success}
For episode $e$, the end-to-end acceptance criterion is
\begin{equation}
Y_e=B_{w(e)}\wedge D_e\wedge M_e\wedge K_e\wedge T_e\wedge
S_e\wedge I_e,
\label{eq:simcode4d-episode-success}
\end{equation}
where $D_e$ denotes overlay validity, $M_e$ mechanics/contact admissibility,
$K_e$ kinematic consistency, $T_e$ the complete task predicate, $S_e$ safety,
and $I_e$ integrity. The primary episode metric is
\begin{equation}
\mathrm{RobustTaskSuccess}
=\frac{1}{N_E}\sum_{e=1}^{N_E}\indicator{Y_e}.
\label{eq:simcode4d-robust-success}
\end{equation}
The seven factors in $Y_e$ are grouped only for presentation:
\emph{Task} denotes $T_e$, while \emph{Physical} denotes
$M_e\land K_e\land S_e$. Build and task-overlay validity remain
$B_{w(e)}$ and $D_e$, and $I_e$ denotes integrity. Thus, the RTS
reported in tables uses the complete acceptance gate above. Build,
overlay, load, timeout, non-finite-state, safety, and integrity
failures remain in the registered denominator as zeros. The five
fixed evaluation physics seeds ($401$, $409$, $419$, $421$, and $431$)
perturb case-declared movable-body position, yaw, and friction; mass
is additionally perturbed when a task contains an independently
movable object. We report per-predicate success,
all-five-seeds pass, forbidden-contact and joint-limit rates, event-order
accuracy, terminal margin, dynamic penetration, and friction-contact coverage.

\subsection{Dev15 Benchmark Validation}

We evaluate the frozen \textsc{Dev15} packages with Python~3.10.8,
MuJoCo~3.9.0, and NumPy~1.26.3. The registered validation matrix contains
15 cases, three deterministic rebuild replicates, and five physics seeds, for 225
executions. Aggregation over unique case--physics pairs yields 75 unique
trials, while the full matrix verifies rebuild determinism and artifact
accounting. Table~\ref{tab:dev15-main} reports implementation readiness and
execution results with explicit denominators.

\begin{table*}[t]
\centering
\caption{Implementation and execution results on the frozen
\textsc{SimCode4D-Dev15-RC1} suite. The registered matrix comprises 225
executions over 75 unique case--physics pairs.}
\label{tab:dev15-main}
\small
\setlength{\tabcolsep}{4pt}
\renewcommand{\arraystretch}{1.08}
\begin{tabularx}{\textwidth}{
    @{}
    >{\raggedright\arraybackslash}p{0.28\textwidth}
    >{\centering\arraybackslash}p{0.13\textwidth}
    >{\raggedright\arraybackslash}X
    @{}
}
\toprule
Measure & Result & Evidence and acceptance criterion \\
\midrule
\multicolumn{3}{@{}l}{\textbf{Artifact readiness}} \\[-1pt]
Registered cases with complete contract quartets
    & 15/15
    & Three cases per domain; schema and cross-references are valid \\
Archived and verified authoring source scenes
    & 12/15
    & Compiled M1 worlds and source pointers are retained for all 15 cases \\
Independently loadable M1 worlds
    & 15/15
    & All M1 world endpoints load under the frozen MuJoCo build \\
Executable runtime packages
    & 15/15
    & All task-runtime endpoints load and execute \\
Case-specific probe--validator pairs
    & 15/15
    & Unique hash-bound paths are frozen for every case \\
Complete physical profiles
    & 15/15
    & Mass and inertia; joints and actuation; collision and contacts;
      friction and solver settings \\
\midrule
\multicolumn{3}{@{}l}{\textbf{Benchmark execution}} \\[-1pt]
Completed executions
    & 225/225
    & Zero runtime errors; 75 unique case--physics trials \\
RobustTaskSuccess
    & 225/225
    & All seven factors of $Y_e$ pass, including build and overlay \\
All five physics seeds pass
    & 15/15
    & Every case passes every registered physics seed \\
Forbidden-contact episodes
    & 0/225
    & Hard safety gate; zero violations required \\
Joint-limit episodes
    & 0/225
    & Hard safety gate; zero violations required \\
Event-order accuracy
    & 210/210
    & Ordered event graphs are defined for 210 executions \\
Friction-cone failures
    & 0/225
    & Failure requires utilization $>1.25$ for at least $10\,\mathrm{ms}$ \\
Cases with friction-contact evidence
    & 14/15
    & 10,160,604 contact samples; the weld case is intentionally non-contact \\
Maximum penetration
    & $0.922\,\mathrm{mm}$
    & Acceptance gate: $\leq 1\,\mathrm{mm}$ \\
Maximum forward/inverse generalized-force residual
    & $2.03\times10^{-9}$
    & Acceptance gate: $\leq 10^{-5}$ \\
\bottomrule\\[-12pt]
\end{tabularx}
\end{table*}

Artifact coverage and physical diagnostics are summarized in
Fig.~\ref{fig:simcode4d-dev15-summary}; panels (c) and (d) report
lower-is-better physical self-consistency measures.

All reference packages pass build and overlay preflight, and all 225
executions pass mechanics, kinematics, task, safety, and integrity
checks. Maximum penetration and generalized-force residual are
$0.921837\,\mathrm{mm}$ and $2.03424\times10^{-9}$, respectively,
both below their gates. Fourteen cases produce friction-bearing
contact; the articulated weld-path case is intentionally non-contact.
Event order passes all 210 eligible executions. These results
establish benchmark readiness under the declared physical profiles.

\section{Experiments}
\label{sec:experiments}

\subsection{Setup}

We distinguish benchmark validation from method evaluation. The frozen
\textsc{Dev15} suite contains 15 cases across five domains and three
task families. Its reference packages are tested with the five
evaluation physics seeds defined in Sec.~\ref{sec:benchmark}, yielding
75 unique case--physics trials. Three deterministic rebuild replicates
produce 225 benchmark-validation executions; neither count is a
generated-method score.

Our source-conditioned experiment uses the 12 cases with verified
authoring sources. Three generation seeds ($1103$, $2207$, and $3301$)
produce 36 world--task packages; these seeds do not index physics
perturbations. We compare default, physics-aware, and repaired outputs
under the five evaluation physics seeds, held out from repair, yielding
180 registered rollouts per condition. Repair observes only the separate
feedback physics seed $503$ and permits at most three execution repairs
and two state refinements. The held-out seeds perturb movable-body pose, yaw, friction,
and mass where applicable.

The matched-prompt comparison uses three prespecified indoor tasks:
living-room pushing, dining-room transport, and kitchen pushing.
Code2Worlds and SceneSmith each generate one world per prompt.
Code2Worlds is scored through a uniform MuJoCo adapter and SceneSmith
through its official export. Code2Worlds-2 (Ours) applies M1/M2 to
the same final Code2Worlds visual output; it is a paired extension,
not another independent text-to-scene generation. Five evaluation
physics seeds yield 15 rollouts per condition with common probes and
source-relative task predicates. This cohort remains separate from
the fixed-source experiment.

Build is scored per package; Task, Physical, and RTS per rollout.
For fixed-source rows, Build scores first outputs except in the
repaired row, which scores the selected post-repair package.
Repairs denotes mean feedback calls per package; a dash marks a
non-comparable native pipeline. RTS follows the complete seven-factor
gate in Sec.~\ref{sec:benchmark}, with failed builds and executions
retained in the denominators. Evaluation physics seeds retest a world,
not independent generations. Scores use headless MuJoCo~3.9.0;
Blender EEVEE serves qualitative visualization only. Fixed-source
runs use a $1\,\mathrm{ms}$ step, \texttt{implicitfast}, the Newton
solver, and gravity $(0,0,-9.81)\,\mathrm{m\,s^{-2}}$; native exports
retain their own simulator options.

\begin{table*}[t]
\centering
\caption{Generated-world acceptance under two non-pooled protocols.}
\label{tab:dev15-exp-physics}
\footnotesize
\setlength{\tabcolsep}{4pt}
\renewcommand{\arraystretch}{1.08}
\begin{tabular*}{\textwidth}{@{\extracolsep{\fill}}lccccc@{}}
\toprule
Method / stage & Build & Task & Physical & RTS & Repairs \\
\midrule
\multicolumn{6}{@{}l}{\emph{Fixed-source progression}} \\
Default export
    & 33/36 & 63/180 & 75/180 & 45/180 & 0 \\
$+$ SimReady scene compilation
    & 33/36 & 62/180 & 93/180 & 50/180 & 0 \\
$+$ SimReady scene compilation $+$ feedback repair
    & 34/36 & 69/180 & 108/180 & \textbf{60/180} & 1.61 \\
\midrule
\multicolumn{6}{@{}l}{\emph{Matched-prompt comparison}} \\
Code2Worlds~\cite{code2worlds}
    & 3/3 & 5/15 & 0/15 & 0/15 & -- \\
\textbf{Code2Worlds-2 (Ours)}
    & \textbf{3/3} & \textbf{15/15} & \textbf{15/15}
    & \textbf{15/15} & \textbf{0} \\
SceneSmith~\cite{scenesmith}
    & 3/3 & 0/15 & 13/15 & 0/15 & -- \\
\bottomrule
\end{tabular*}
\end{table*}

\subsection{Generated-World Compatibility and Robust Success}

Table~\ref{tab:dev15-exp-physics} reports generated-world results under two
separate protocols; the \textsc{Dev15} reference-package checks are reported in
Sec.~\ref{sec:benchmark}. In the fixed-source experiment, physical acceptance
increases from $75/180$ for the default export to $93/180$ after physical
enrichment and $108/180$ after repair. RTS increases from $45/180$ to $50/180$
and $60/180$, respectively. Task success does not improve at every stage
($63/180$ to $62/180$ after enrichment). The raw-to-repaired RTS gain is
$8.3$ percentage points; its source-case-clustered 95\% bootstrap interval is
$[-11.1,\,28.9]$ percentage points.

For the 36 first generated fixed-source plans,
\texttt{TaskOverlayPass@1} is 36/36, and required-entity binding
micro-F1 is 97.3\% (108 TP, 3 FP, 3 FN). Initial task-object
support-relation micro-F1 is 66.7\% (15 TP, 15 FP, 0 FN), evaluated
only on five annotated cases across three generation seeds. These
coding diagnostics measure overlay and semantic bindings, not
simulator-state task success.

Under the three matched prompts, the uniform MuJoCo adapter of Code2Worlds
passes $5/15$ task trials but no physical trials. Applying M1/M2 to the same
Code2Worlds visual outputs yields $15/15$ for task, physical, and RTS
acceptance; all three packages pass before feedback repair. SceneSmith's
official MuJoCo exports pass $13/15$ physical trials and no task trials.
These are descriptive outcomes for one generated world per prompt, not a
population-level ranking.

The physical gate applies the frozen mechanics, kinematics, collision, and
friction checks in Sec.~\ref{sec:benchmark}, including the $1\,\mathrm{mm}$
penetration limit and sustained friction-cone criterion.
Fig.~\ref{fig:simready-export-comparison} illustrates the geometric mechanism:
a single convex hull can occlude an opening that decomposed colliders preserve.
All quantitative results come from simulator-state rollouts rather than the
illustration.

\begin{figure*}[t]
\centering
\includegraphics[width=0.99\textwidth]
{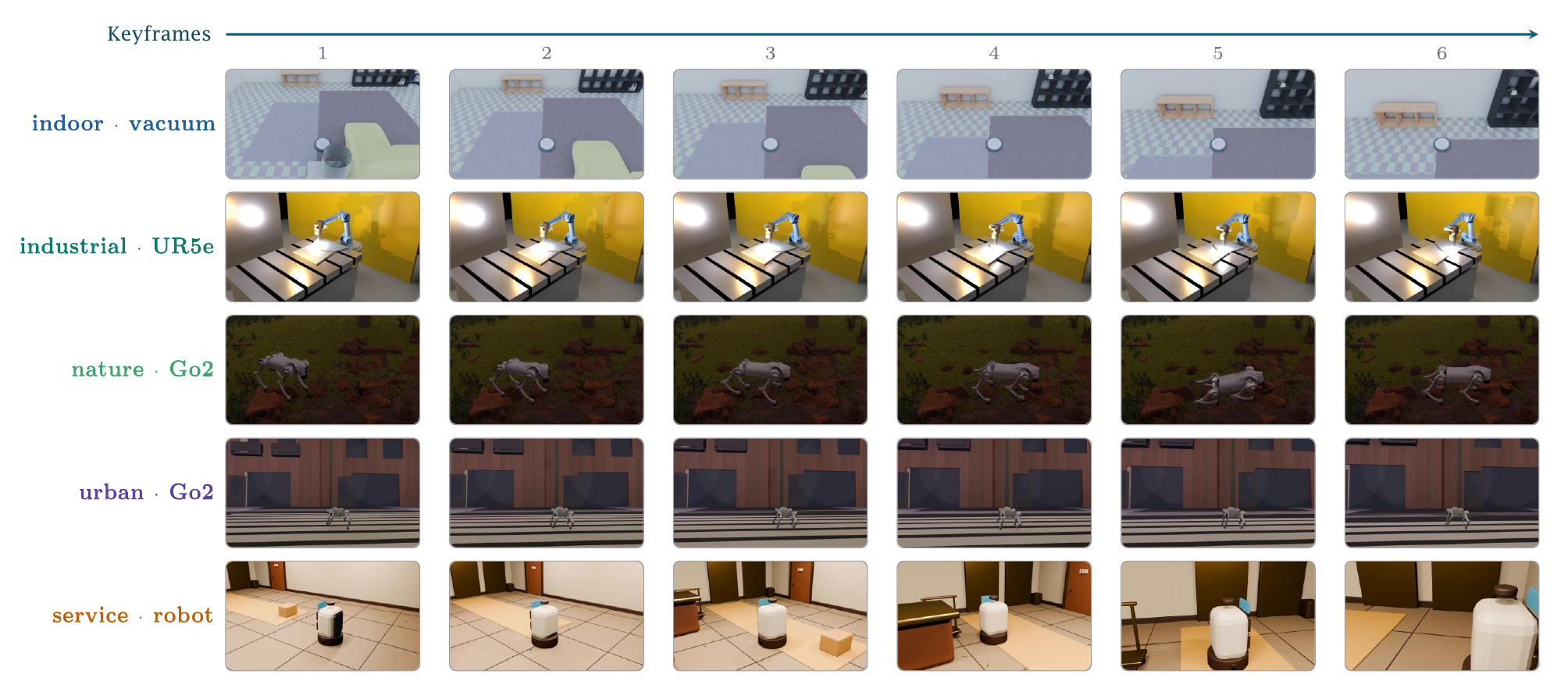}
\caption{Representative task rollouts from the five evaluation domains. Rows
show one case per domain, and columns show six temporally ordered keyframes.}
\label{fig:qualitative-rollouts}
\end{figure*}

\subsection{Embodied Task Execution}

Task outcomes are determined by frozen simulator-state predicates, not by
visual inspection. Navigation requires compliant, collision-free goal arrival
and stopping; pick-and-transport requires a legal grasp, transfer, and terminal
placement; dynamic-interaction tasks require their case-specific ordered
events.

Table~\ref{tab:dev15-exp-tasks} breaks down the 12-case fixed-source results
by task family. After repair, the task predicates pass in $35/60$ navigation,
$15/30$ pick-and-transport, and $19/90$ dynamic-interaction trials; the
corresponding RTS counts are $35/60$, $14/30$, and $11/90$. The aggregate RTS
gain is therefore not uniform: navigation decreases from $45/60$ for the
default export to $35/60$, whereas pick-and-transport and dynamic interaction
increase from zero to $14/30$ and $11/90$, respectively. The matched-prompt
methods are reported separately in Table~\ref{tab:dev15-exp-physics} and are
not pooled into these task-family totals.

Fig.~\ref{fig:qualitative-rollouts} shows temporal progression in five frozen
\textsc{Dev15} \emph{reference-package} rollouts. These frames illustrate the
task definitions; they are neither method-generated examples nor inputs to
the state-based scores.

\begin{table}[t]
\centering
\caption{Task-family RTS for the 12 fixed-source cases. Each case contributes
three generated packages and five evaluation physics seeds.}
\label{tab:dev15-exp-tasks}
\footnotesize
\renewcommand{\arraystretch}{1.08}
\begin{tabular*}{\columnwidth}{@{\extracolsep{\fill}}lcccc@{}}
\toprule
Task family & Cases & Default & Enriched & Repaired \\
\midrule
Navigation     & 4 & 45/60 & 35/60 & 35/60 \\
Pick/transport & 2 & 0/30  & 14/30 & 14/30 \\
Dynamic        & 6 & 0/90  & 1/90  & 11/90 \\
\bottomrule
\end{tabular*}
\end{table}

\subsection{Ablations}
\label{sec:ablations}

We evaluate three physical-preparation interventions and four refinement
feedback rules. Within each paired comparison, the task specification, robot
policy, simulator settings, validator, and perturbation seeds remain fixed.
Eligibility differs across interventions; their denominators are therefore
reported separately rather than pooled. Table~\ref{tab:simcode4d-ablations}
reports the Module~1 interventions, and
Table~\ref{tab:simcode4d-critic-loop} compares Module~2 feedback rules from
the same 36 enriched starting worlds.

\begin{table}[t]
\centering
\caption{Module~1 ablations on paired eligible inputs. The first two
interventions report held-out physics trials; pose stabilization reports
initial object states. Higher is better for every measure.}
\label{tab:simcode4d-ablations}
\footnotesize
\setlength{\tabcolsep}{3pt}
\begin{tabular*}{\columnwidth}{@{\extracolsep{\fill}}llcc@{}}
\toprule
Ablated variant & Measure & Full & Ablated \\
\midrule
Uniform attributes & RTS & 69/75 & 54/75 \\
                   & Physical pass & 69/75 & 55/75 \\
\addlinespace[2pt]
Single convex hull & RTS & 15/15 & 10/15 \\
                   & Physical pass & 15/15 & 11/15 \\
\addlinespace[2pt]
No pose stabilization & Stable & 8/12 & 1/12 \\
                      & Penetration-safe & 11/12 & 0/12 \\
\bottomrule
\end{tabular*}
\end{table}

\begin{table}[t]
\centering
\caption{Module~2 feedback ablation from 36 matched enriched starts.
Build is measured over 36 worlds; task, physical, and RTS outcomes are
measured over 180 held-out physics trials. Failed builds remain in the
denominator.}
\label{tab:simcode4d-critic-loop}
\footnotesize
\setlength{\tabcolsep}{3pt}
\begin{tabular*}{\columnwidth}{@{\extracolsep{\fill}}lcccc@{}}
\toprule
Feedback rule & Build & Task & Physical & RTS \\
\midrule
No repair & 33/36 & 62/180 & 93/180 & 50/180 \\
Visual only & 36/36 & 84/180 & 112/180 & 67/180 \\
State only & 36/36 & 96/180 & 116/180 & \textbf{81/180} \\
Fused, fixed budget & 35/36 & 81/180 & 99/180 & 62/180 \\
\bottomrule
\end{tabular*}
\end{table}

\paragraph*{Physical attributes}
Five eligible packages are evaluated under 15 paired physics perturbations
each. The ablation replaces task-object mass, diagonal inertia, and sliding,
torsional, and rolling friction with uniform defaults; robot hardware and
control remain unchanged. Object-specific attributes yield 69/75 RTS and
69/75 physical passes, compared with 54/75 and 55/75 under uniform defaults.
For RTS, 19 trial pairs favor object-specific attributes, four favor the
defaults, and 52 tie. Much of the difference occurs in medicine transport,
where the defaults cause joint-limit violations. This intervention tests
sensitivity to assigned properties, not the accuracy of VLM estimates.

\paragraph*{Collision decomposition}
The collision ablation covers one stone and two cups, each under five paired
physics seeds. Replacing their CoACD~\cite{coacd} colliders with a single
convex hull, while retaining other physical parameters, reduces RTS from
15/15 to 10/15 and physical acceptance from 15/15 to 11/15. All five RTS
losses occur in the cup-pushing task; the other two tasks retain success.
The observed benefit is therefore task-dependent.

\paragraph*{Pose stabilization}
Four objects are tested from three initial offsets each. The full
preprocessing step applies bounded vertical separation followed by a 5-s
gravity pass; the ablation omits both operations. A separate 5-s validation
increases the stable-rest count from 1/12 to 8/12 and the count satisfying
the 1-mm penetration gate from 0/12 to 11/12. These are object-level
rest-state diagnostics, not task-success rates. A contact audit also shows
that a stable rest pose does not necessarily preserve the intended support
relation. Because separation and settling are removed together, this
experiment does not isolate either operation or the support graph individually.

\paragraph*{Critic feedback}
Table~\ref{tab:simcode4d-critic-loop} compares no repair, visual-only
feedback, simulator-state-only feedback, and their fusion under matched
limits of three execution repairs and two state refinements. Feedback uses physics seed $503$; final outcomes use the five
evaluation physics seeds per world. The state-only critic includes task predicates as well as physical
diagnostics and should not be described as physics-only. Under this
fixed-budget protocol, state-only feedback achieves the highest RTS
(81/180). Visual-only feedback reaches 67/180, fusion reaches 62/180,
and no repair reaches 50/180. In 23/36 fused runs, compile-invalid
proposals exhaust the execution-repair budget. This fixed-budget fusion
experiment does not use the main experiment's early-accept rule, so its
62/180 result is distinct from the repaired result in the main table.

A separate blinded acceptance diagnostic contains 36 candidates:
19 satisfy the registered simulator predicates and 17 do not. Visual-only
judgment accepts 3/17 invalid candidates, whereas adding the
simulator-state gate accepts 0/17; both rules accept 6/19 valid candidates.
Zero invalid acceptance follows from the gate's definition. This
diagnostic measures candidate filtering, not closed-loop repair quality.

Finally, 34 matched initial--final repair trajectories improve from
50/170 to 60/170 RTS trials: four worlds improve, one regresses, and
29 are unchanged. Intermediate rounds have different eligible cohorts;
these endpoint results do not imply monotonic improvement at every round.

\section{Conclusion}
We presented Code2Worlds-2, an extension of Code2Worlds that turns 4D world generation into embodied world simulation by appending two modules to the original pipeline and orchestrating them with an agent harness: a SimReady Scene Compiler that endows generated scenes with collision geometry, physical attributes, a support graph, and physical constraints, and applies a gravity settling pass, and a Simulator-in-the-Loop 4D Refiner that closes a compile--simulate--critique--refine loop to compile visual dynamics into executable rigid and articulated dynamics and validate them through simulator signals rather than pixels alone. We further proposed SimCode4D, a state-based benchmark with five task templates for evaluating embodied simulation. Across the fixed-source cases and the matched prompts, state-based evaluation shows that physical acceptance rises from $75/180$ to $108/180$ and RTS from $45/180$ to $60/180$ after enrichment and repair. Our framework currently covers a limited set of articulated objects and leaves high-fidelity indoor wall collision and general grasping as future work; extending the compiler to Isaac Sim/USD export and a larger manipulation suite are natural next steps.

\ifCLASSOPTIONcompsoc
  \section*{Acknowledgments}
\else
  \section*{Acknowledgment}
\fi

This work is supported by the Fundamental Research Funds for the Central Universities, Peking University.

\ifCLASSOPTIONcaptionsoff
  \newpage
\fi

\bibliographystyle{IEEEtran}
\bibliography{main}

\begin{IEEEbiography}[{\includegraphics[width=1in,height=1.25in,clip,keepaspectratio]{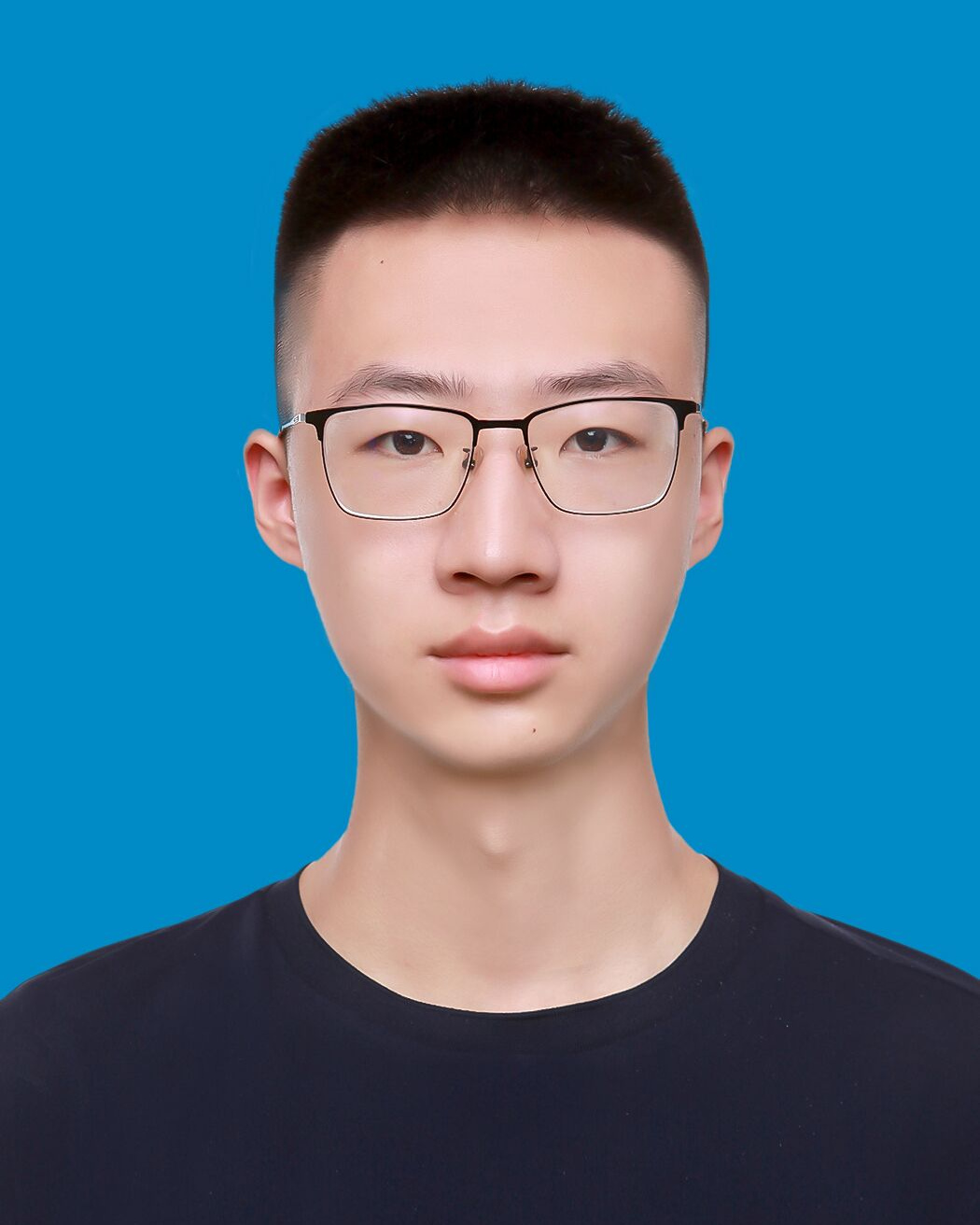}}]{Haoyu Zhang}
is an undergraduate student at the School of Information and Software Engineering, University of
Electronic Science and Technology of China, Chengdu, China, and is currently a visiting student at
Peking University, Beijing, China. His research interests include remote-sensing image fusion
(pansharpening), embodied world simulation, 3D/4D scene generation, and physics-based simulation.
\end{IEEEbiography}

\begin{IEEEbiography}[{\includegraphics[width=1in,height=1.25in,clip,keepaspectratio]{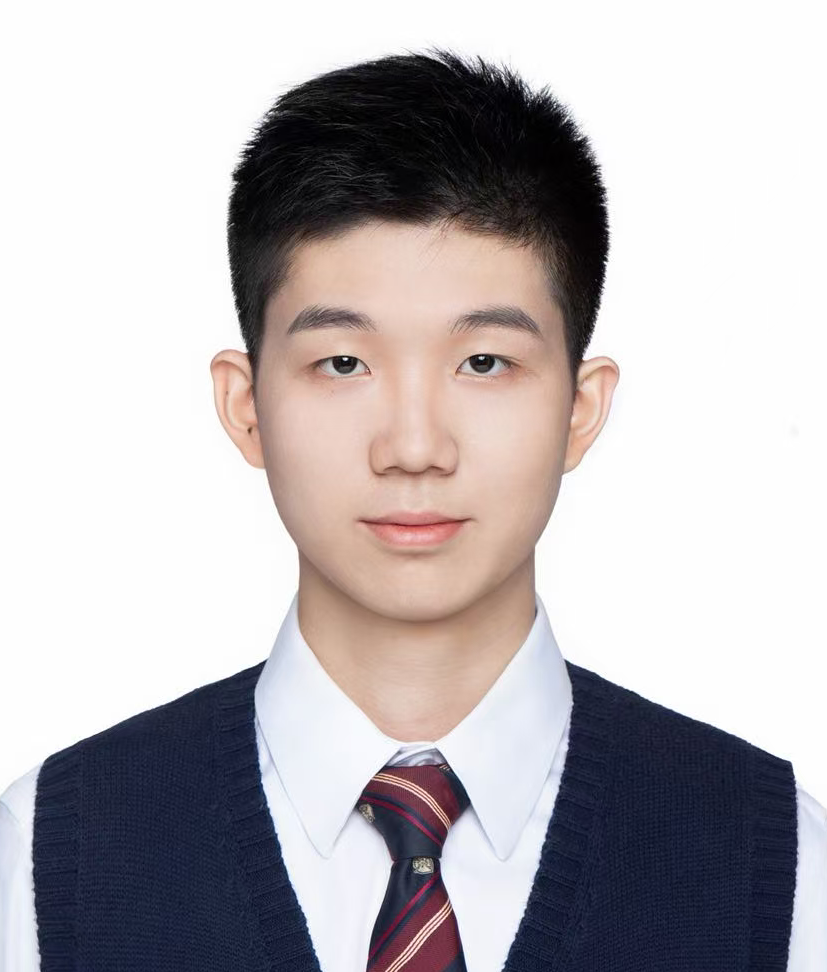}}]{Siming Li}
is currently pursuing his undergraduate degree in the School of Telecommunications Engineering at Xidian University, Xi'an, China. His research interests focus on visual scene parsing and segmentation, embodied AI, physics-based interactive world modeling, and 3D generative tasks such as 3D Gaussian Splatting (3DGS).
\end{IEEEbiography}

\begin{IEEEbiography}[{\includegraphics[width=1in,height=1.25in,clip,keepaspectratio]{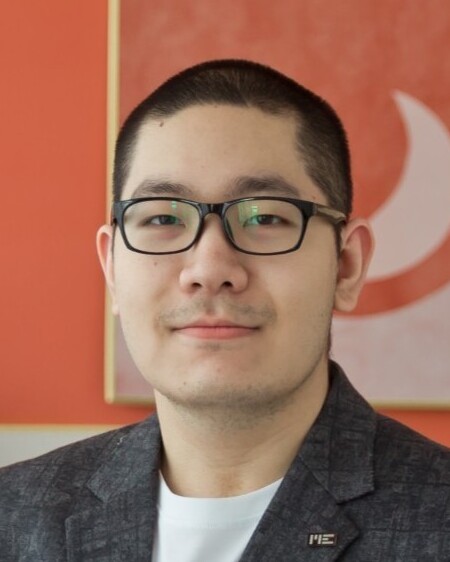}}]{Zeyu Zhang}
is a researcher working on generative AI, with a particular interest in building models that understand and interact with the physical world. He received his bachelor's degree from the Australian National University, where he was advised by Prof. Richard Hartley and Prof. Ian Reid. His research explores generative modeling for learning physical dynamics from visual data. His work spans world models, multimodal foundation models, embodied AI, and AI for health.
\end{IEEEbiography}

\begin{IEEEbiography}[{\includegraphics[width=1in,height=1.25in,clip,keepaspectratio]{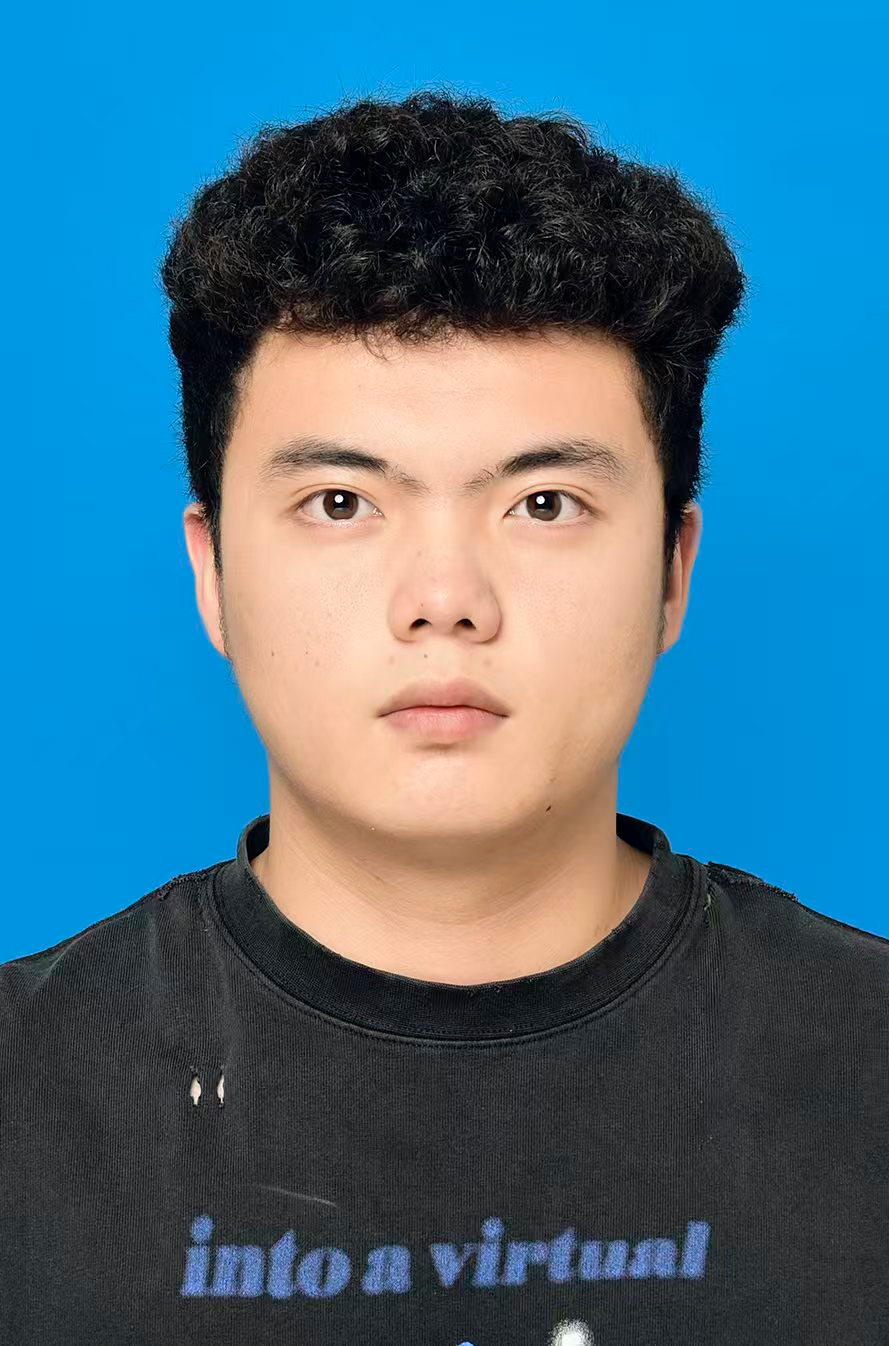}}]{Yi Zhang}
is an undergraduate student majoring in Software Engineering at Southwest University, Chongqing, China. His research interests include video generation, physically grounded world models, 3D/4D vision, multimodal large language models and agents, and physics-aware simulation.
\end{IEEEbiography}

\begin{IEEEbiography}[{\includegraphics[width=1in,height=1.25in,clip,keepaspectratio]{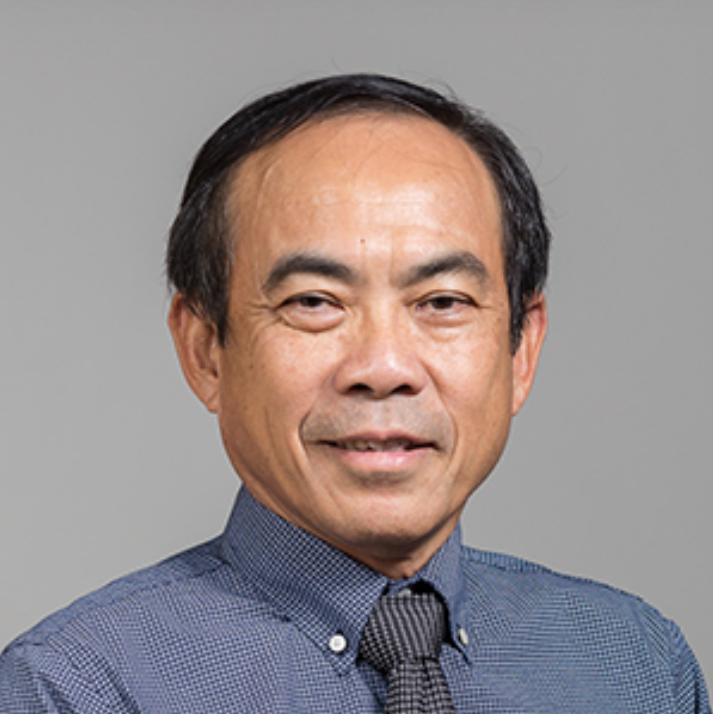}}]{Tat-Seng Chua} received the Ph.D. degree from the
University of Leeds, U.K. He is the KITHCT chair professor with the School of Computing, National University of Singapore, where he was the acting and founding dean of the School from 1998 to 2000. He is
the co-director of NExT, a joint center between NUS and Tsinghua University, to develop technologies for live social media search. He is the 2015 winner of
the prestigious ACM SIGMM Award.
\end{IEEEbiography}

\begin{IEEEbiography}[{\includegraphics[width=1in,height=1.25in,clip,keepaspectratio]{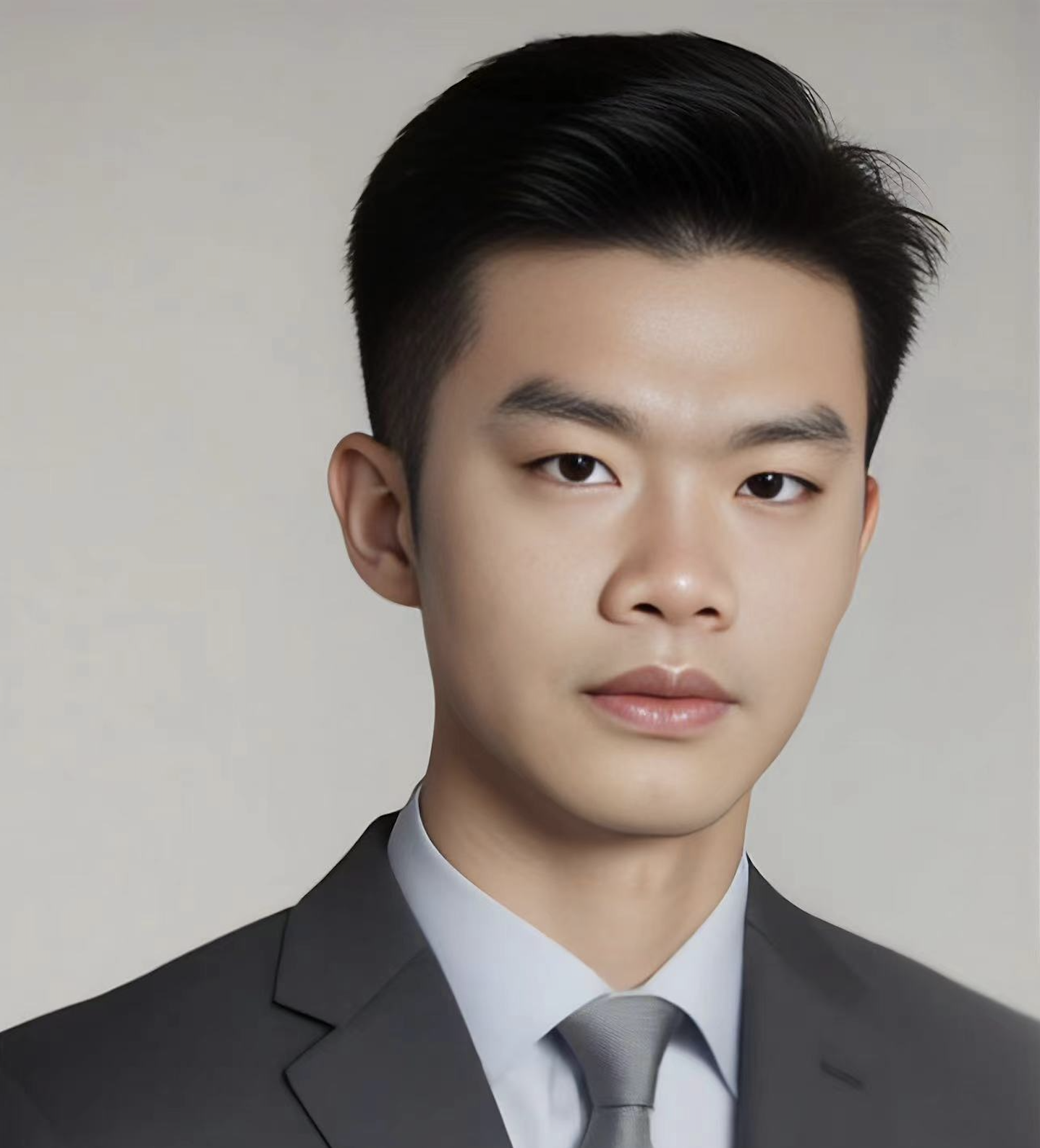}}]{Hao Tang}
is an Assistant Professor at Peking University, China. Previously, he held postdoctoral positions at CMU, USA, and ETH Z\"{u}rich, Switzerland. He earned his master's degree from Peking University, and his Ph.D. from the University of Trento, Italy.
He has had the opportunity to visit the University of Oxford, NEU, NUS, and IIAI, among other institutions.
His research focuses on generative AI, world models, spatial intelligence, embodied AI, LLMs, MLLMs, and computer vision.
\end{IEEEbiography}
\vfill

\clearpage
\onecolumn
\appendices
\numberwithin{figure}{section}
\numberwithin{table}{section}

\section{Demo Screenshots}
\label{app:demos}
\label{app:firstpage}

This section collects full-resolution screenshots from the delivered runs of all
fifteen cases. Each figure covers one domain: every row is one rollout, ordered
top to bottom as in Table~\ref{tab:app-case-index}, and the three columns show
temporally ordered stages (early, middle, late) of that rollout. The frames are
rendered with the pipeline's Blender rendering path and are shown for qualitative
inspection only.

\begin{center}
\includegraphics[width=0.86\linewidth]{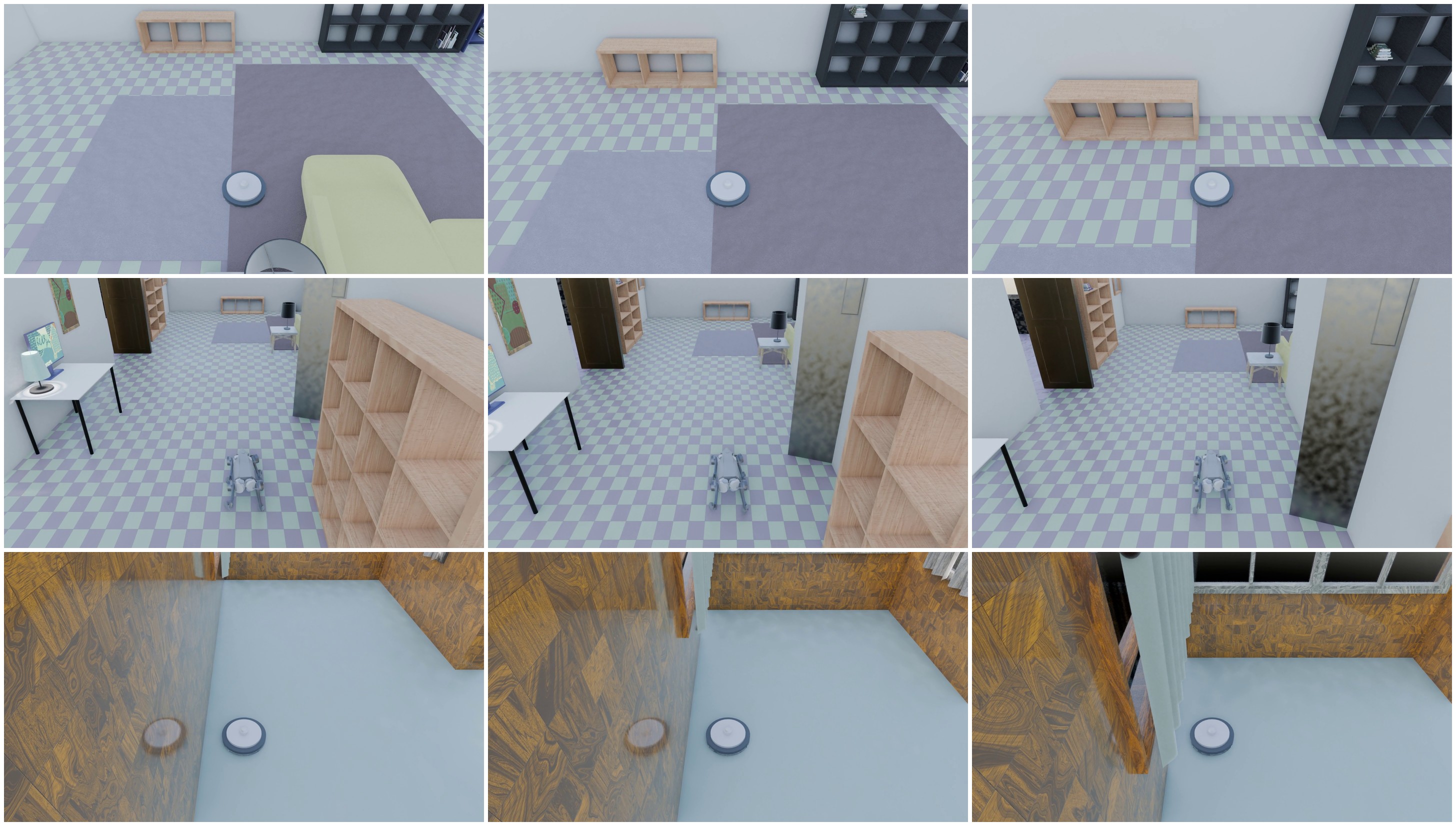}
\end{center}

\captionof{figure}{Indoor (obstacle navigation). Rows: the mobile vacuum driving through the generated living room, the Unitree Go2 traversing the living room, and the mobile vacuum crossing the west corridor. The corridor, rug, sofa, shelving, and sideboard are produced by the scene-coding stage, and each robot is executed in the compiled world.}
\label{fig:app-demo-indoor}

\begin{center}
\includegraphics[width=0.86\linewidth]{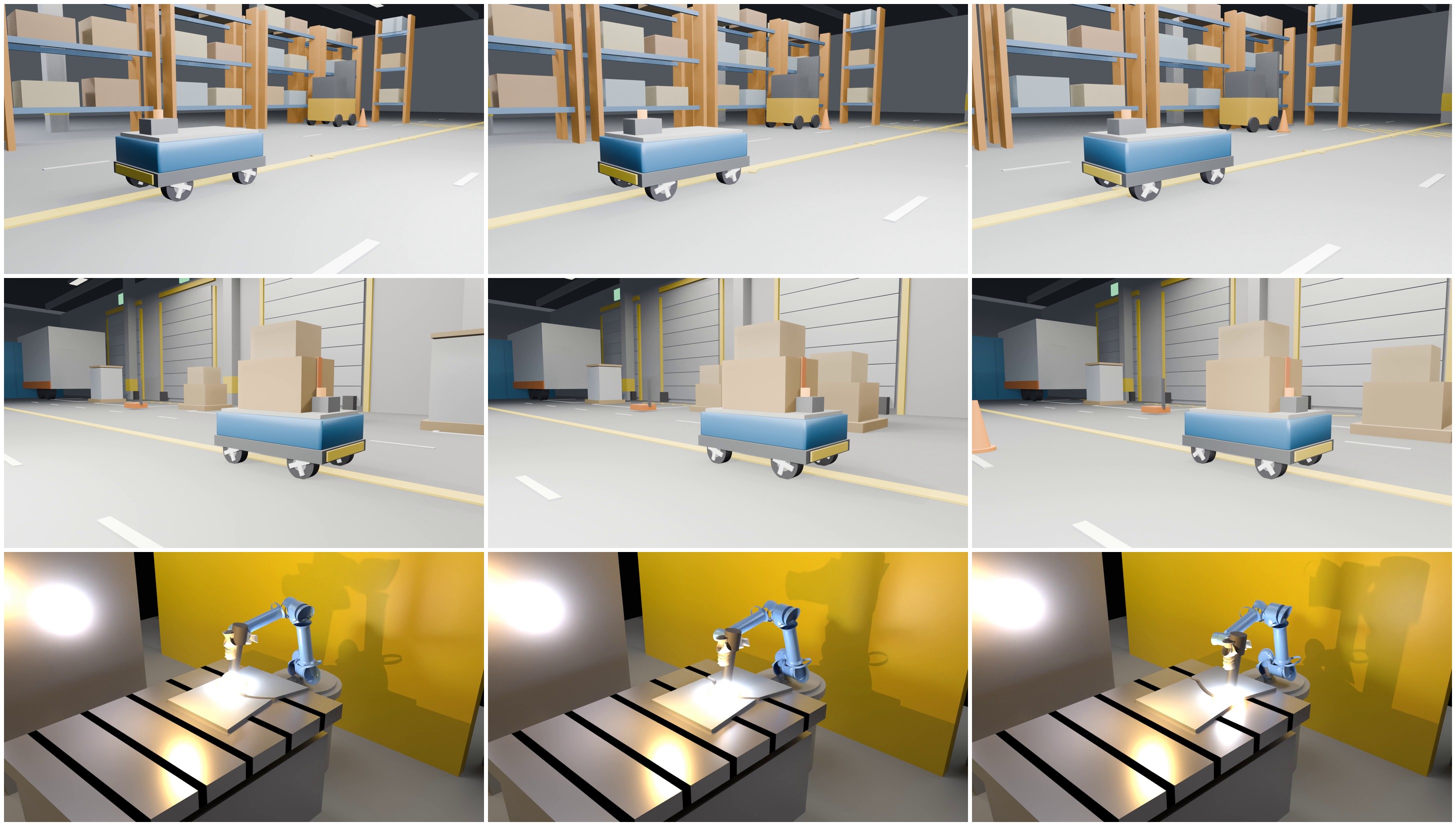}
\end{center}

\captionof{figure}{Industrial. Rows: the four-wheel AGV holding the aisle lane, the AGV transporting its welded cargo, and the UR5e tracing the frozen weld seam on the slotted workstation. Each rollout is executed in the compiled world; the weld tool path is an articulated trajectory with joint limits and contact events validated by the evaluator.}
\label{fig:app-demo-industrial}

\begin{center}
\includegraphics[width=0.86\linewidth]{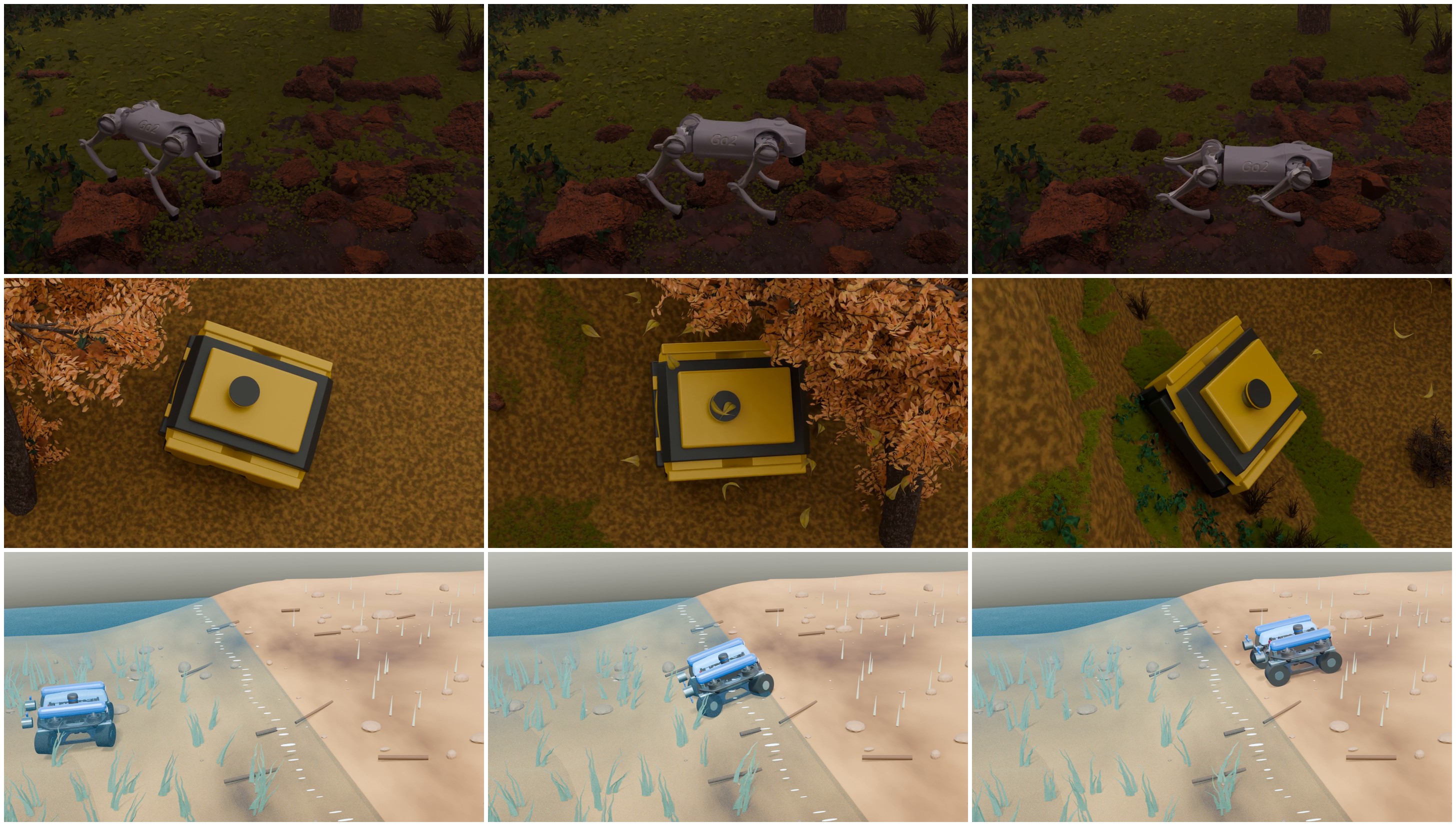}
\end{center}

\captionof{figure}{Natural outdoor. Rows: the Unitree Go2 disturbing the source-bound rock on the generated canyon terrain, the Clearpath Jackal driving the authored route across the hills, and the amphibious rover moving from the water field onto dry land. The non-convex height field is sliced into per-cell convex prisms so that contact follows the local surface. The canyon, hill, and shore terrains are generated scene-code assets; the rock, the route obstacles, and the water field are frozen benchmark entities carrying their own physics-seed perturbations, and all three rollouts are rendered at 1080p from the same compiled worlds that the evaluator scores. The canyon wall, the route markers, and the shore bank are compiled colliders, so the robots' contacts with them are resolved by the same collision pipeline as in the indoor and urban suites. The rows correspond to Natural outdoor-1 (rock kick), -2 (hills turn stop), and -3 (water land) of the frozen index; the rock's terminal rest, the Jackal's stop inside the goal region, and the rover's arrival on dry land are the terminal predicates checked by the case validators. Frames are sampled uniformly across each rollout (early, middle, late).}
\label{fig:app-demo-nature}

\begin{center}
\includegraphics[width=0.86\linewidth]{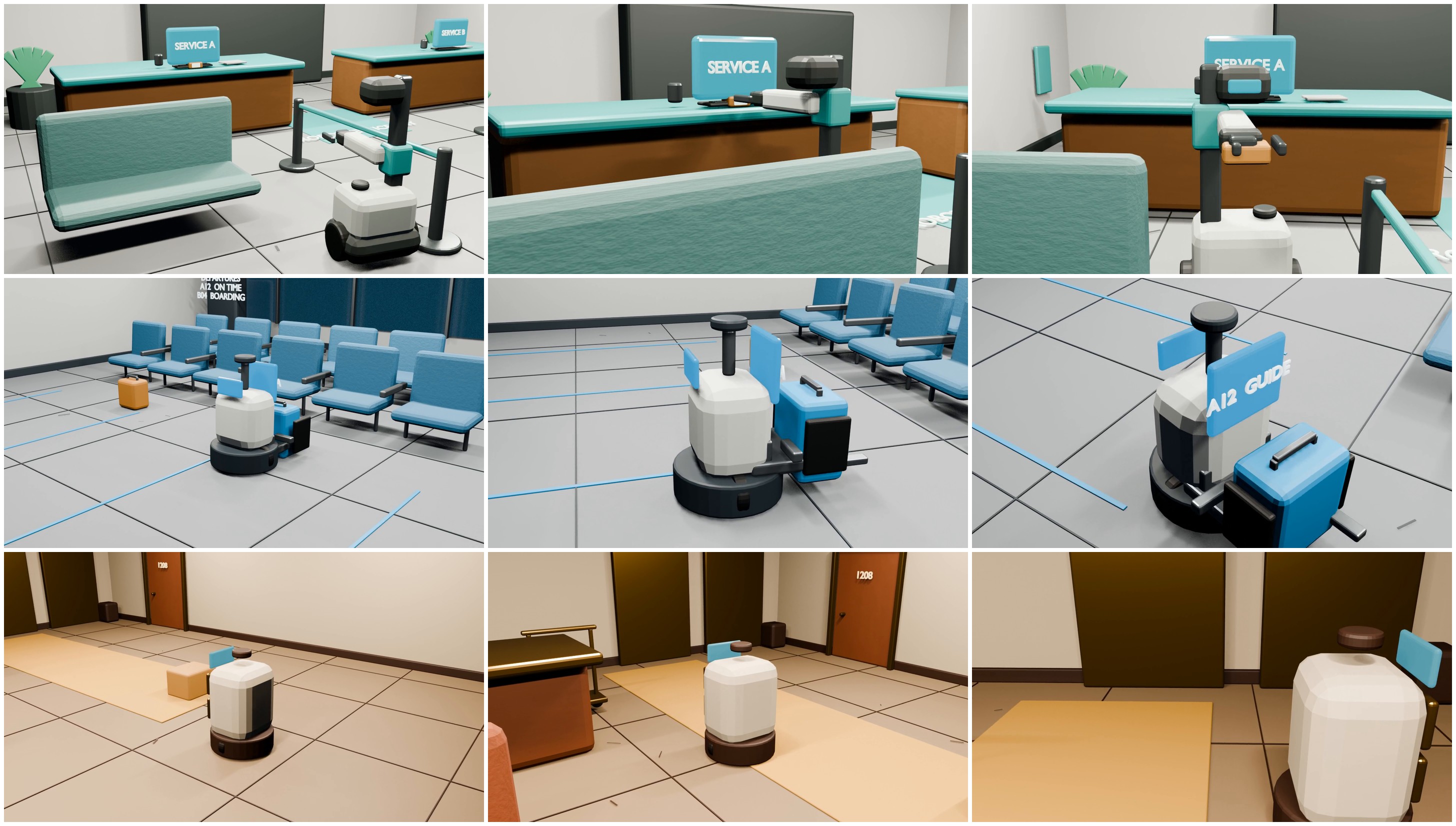}
\end{center}

\captionof{figure}{Service. Rows: the medicine robot delivering the pouch, the airport luggage robot lifting the blue luggage assembly, and the hotel robot opening its payload drawer so that the meal parcel stays supported. The clinic, airport, and hotel interiors are generated scene-code assets with their furniture layouts; the pouches, luggage, and meal parcel are dynamic bodies whose grasps are issued through the benchmark grasp API, and the grasp and terminal-state predicates are checked by the case validator. The rows correspond to Service-1 (medicine), -2 (airport luggage), and -3 (hotel drawer) of the frozen index; the drawer is opened through the benchmark actuation API, and the retained-grasp and 0.5-second support conditions are the terminal predicates. The three interiors stress the same grasp stack across different geometry: a counter hand-off, a floor-level suitcase assembly, and a drawer payload.}
\label{fig:app-demo-service}

\begin{center}
\includegraphics[width=0.86\linewidth]{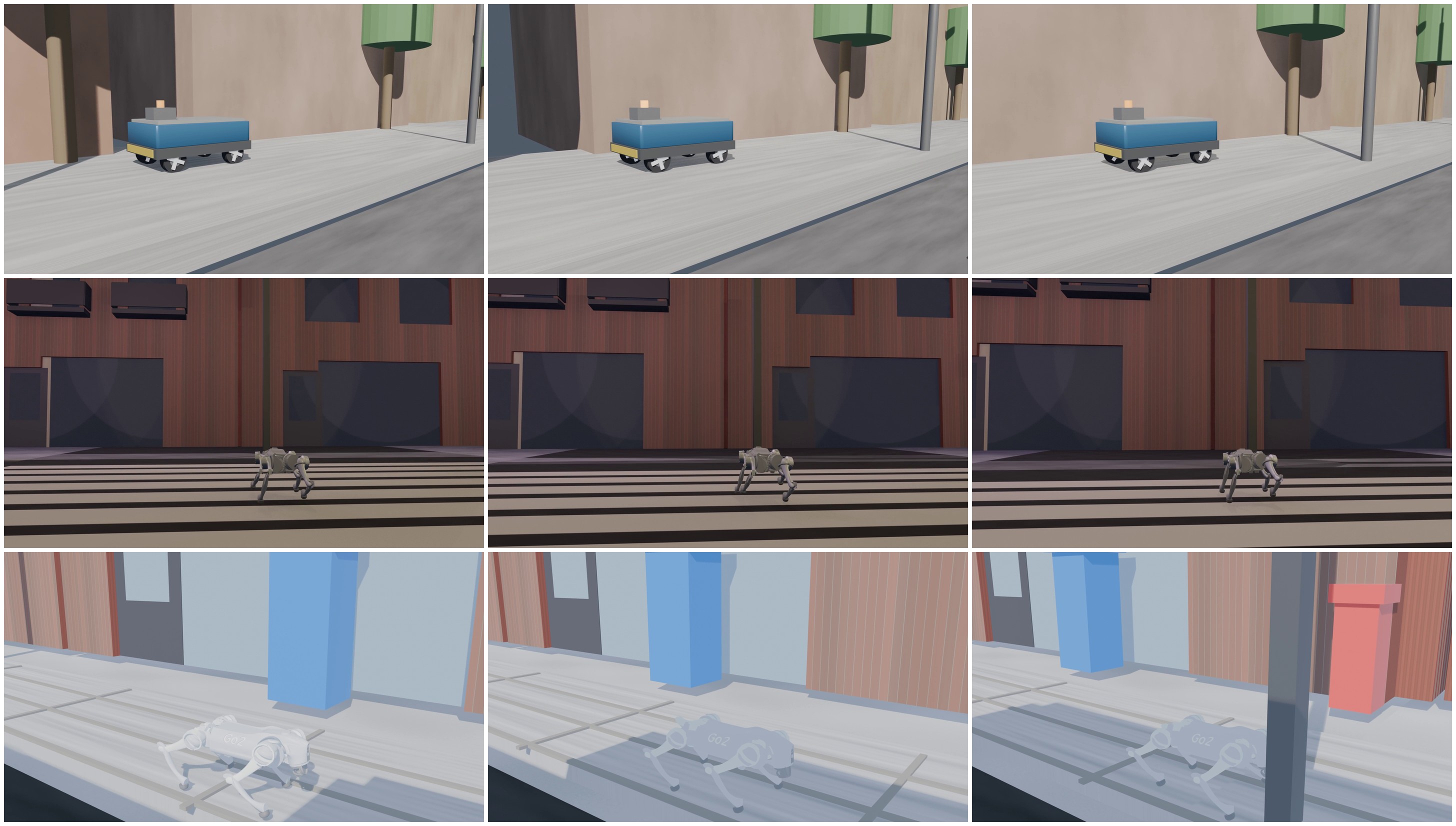}
\end{center}

\captionof{figure}{Urban. Rows: the four-wheel AGV driving along the city pavement, the Unitree Go2 crossing the road on the generated zebra crossing, and the Go2 making contact with the frozen target pole. The crossing and the pole are compiled collision surfaces, and the corridor constraint is enforced by the evaluator.}
\label{fig:app-demo-urban}

\section{SimCode4D Case Specifications}
\label{app:cases}

\textsc{SimCode4D-Dev15-RC1} contains 15 world--task cases, three in each of five
domains. The realized task families are obstacle navigation (7 cases),
pick-and-transport (2 cases), and dynamic interaction (6 cases). Every case
freezes four machine-readable contracts together with an independent probe and
validator; Table~\ref{tab:app-case-index} lists the case index and
Table~\ref{tab:app-contracts} summarizes the frozen artifacts. These fifteen
frozen reference packages are distinct from the three matched-prompt case
studies of Appendix~\ref{app:simcode4d-matched-prompts}, which are
experiment-specific inputs generated by the compared native pipelines rather
than members of the frozen suite.

{\scriptsize
\setlength{\tabcolsep}{3pt}
\begin{longtable}{@{}>{\raggedright\arraybackslash}p{0.245\linewidth}>{\raggedright\arraybackslash}p{0.105\linewidth}>{\raggedright\arraybackslash}p{0.23\linewidth}>{\raggedright\arraybackslash}p{0.115\linewidth}p{0.25\linewidth}@{}}
\caption{The frozen \textsc{SimCode4D-Dev15-RC1} case index. Cases are listed per domain in the frozen suite order; each row gives the scene, the benchmark-owned robot, the realized task family, and the natural-language goal frozen in \texttt{case\_spec.json}.}
\label{tab:app-case-index}\\
\toprule
Case & Domain & Robot & Task family & Goal (frozen prompt) \\
\midrule
\endfirsthead
\toprule
Case & Domain & Robot & Task family & Goal (frozen prompt) \\
\midrule
\endhead
\bottomrule
\endlastfoot
Indoor-1 \newline \textit{vacuum living room} & Indoor & mobile vacuum & Obstacle nav. & Drive the vacuum east through the living-room corridor, enter the frozen goal, and remain stopped for 0.5 seconds. \\
Indoor-2 \newline \textit{go2 living room} & Indoor & Unitree Go2 quadruped & Obstacle nav. & Traverse the living room east with the Go2, stay in the frozen corridor, enter the goal, and stop. \\
Indoor-3 \newline \textit{vacuum west} & Indoor & mobile vacuum & Obstacle nav. & Drive the vacuum west through the indoor corridor, enter the frozen goal, and remain stopped for 0.5 seconds. \\
Industrial-1 \newline \textit{agv aisle} & Industrial & four-wheel AGV & Obstacle nav. & Drive the four-wheel AGV east along the industrial aisle, hold the lane, enter the goal, and stop. \\
Industrial-2 \newline \textit{agv cargo} & Industrial & four-wheel AGV with welded cargo & Dynamic interaction & Transport the physically modeled cargo east with the AGV, retain the declared weld attachment, enter the cargo goal, and stop. \\
Industrial-3 \newline \textit{weld} & Industrial & UR5e welding arm & Dynamic interaction & Trace the frozen weld seam continuously with the UR5e tool tip and finish at rest. \\
Natural outdoor-1 \newline \textit{rock kick} & Natural outdoor & Unitree Go2 quadruped & Dynamic interaction & Use the Go2 feet to make the source-bound rock move physically in the uneven canyon, then let it settle. \\
Natural outdoor-2 \newline \textit{hills turn stop} & Natural outdoor & Clearpath Jackal UGV & Obstacle nav. & Drive the Jackal across the authored hills, complete the authored left turn, enter the goal region, and remain stopped for 0.5 seconds. \\
Natural outdoor-3 \newline \textit{water land} & Natural outdoor & four-wheel amphibious rover & Dynamic interaction & Move the amphibious rover from the declared static water field, contact the bank, reach dry land, and stop in the land goal. \\
Service-1 \newline \textit{medicine} & Service & differential-drive service robot with lift and arm & Pick \& transport & Approach, legally grasp, and deliver the medicine pouch to the authored delivery region while retaining a stable grasp. \\
Service-2 \newline \textit{airport luggage} & Service & differential-drive airport luggage robot with lift & Pick \& transport & Approach the blue luggage assembly, legally grasp and lift it, then deliver it to the authored airport goal while retaining a stable grasp. \\
Service-3 \newline \textit{hotel drawer} & Service & differential-drive hotel service robot with payload drawer & Dynamic interaction & Navigate the hotel robot to the terminal region, stop, open its payload drawer, and keep the meal parcel supported for 0.5 seconds. \\
Urban-1 \newline \textit{agv city} & Urban & four-wheel AGV & Obstacle nav. & Drive the AGV east along the city pavement, stay in the frozen corridor, enter the goal, and stop. \\
Urban-2 \newline \textit{intersection} & Urban & Unitree Go2 quadruped & Obstacle nav. & Start on the west sidewalk, cross the road east through the frozen corridor, reach the east sidewalk, and stop. \\
Urban-3 \newline \textit{pole blocking} & Urban & Unitree Go2 quadruped & Dynamic interaction & Approach the frozen target pole with the Go2, make physical contact, remain blocked without passing through it, and stop. \\
\end{longtable}
}

{\footnotesize\setlength{\tabcolsep}{4pt}
\begin{center}
\begin{tabularx}{\linewidth}{@{}p{0.235\linewidth}X@{}}
\toprule
Frozen artifact & Content and role \\
\midrule
\texttt{case\_spec.json} & Semantic entities and their required bindings, observation channels (bodies and joints), benchmark robot identity, simulator configuration, probe and validator references, and canonical-source pointers (schema \texttt{simcode4d.rc3a-dev1.case-spec}). \\
\texttt{case\_gold.json} & Evaluator-owned typed predicates (initial, event, always, terminal) with terminal tolerances and the physics-seed perturbation definition. \\
\texttt{physics\_profile.json} & Body mass and inertia, joint type/range/damping, actuator gears and control limits, collision masks, semantic contact groups, and sliding/torsional/rolling friction. \\
\texttt{task\_confirm.yaml} & Author-confirmed robot identity and task statement, the canonical source \texttt{.blend} with byte size and SHA-256, start state, and acceptance summary. \\
probe / validator & One probe script (robot actuation, grasp API, perturbations) and one independent validator per case, hash-bound inside \texttt{case\_spec.json}. \\
\bottomrule
\end{tabularx}
\end{center}
\captionof{table}{Contracts frozen for every case. All four files are present for each of the 15 cases; the probe and validator named in \texttt{case\_spec.json} are owned by the benchmark, so a submitted method cannot encode the evaluation policy in its world program.}
\label{tab:app-contracts}
}

\subsection{Case Contract Fields}
\label{app:case-fields}
The following table summarizes the fields frozen for every case, including the
start state, interaction constraints, acceptance predicates, and perturbation
definition summarized above.

{\footnotesize\setlength{\tabcolsep}{4pt}
\begin{center}
\begin{tabularx}{\linewidth}{@{}>{\raggedright\arraybackslash}p{0.18\linewidth}X@{}}
\toprule
Field & Content frozen in the contracts \\
\midrule
Identity & Case, source world, scene domain, and the canonical source \texttt{.blend} with byte size and SHA-256. \\
Task & Benchmark-owned robot (identity, binding, mobility), the natural-language goal, and the realized task family. \\
Initial state & Free-body poses and joint values of the start state, the required entities and their bindings, and the observation channels (bodies and joints). \\
Interaction & Required event order, permitted contacts and semantic contact groups, and any required attachment (for example a welded cargo). \\
Acceptance & Typed predicates (initial, event, always, terminal) with terminal tolerances, plus the hard safety gates (forbidden contacts, joint limits). \\
Perturbation & The physics-seed perturbations applied to movable bodies (position, yaw, friction; mass where independently movable). \\
Probe and validator & The benchmark probe (actuation, grasp API, perturbations) and an independent validator, both hash-bound. \\
\bottomrule
\end{tabularx}
\end{center}
\captionof{table}{Fields frozen per case. The same structure is instantiated for all 15 cases; the probe and validator are owned by the benchmark.}
\label{tab:app-case-fields}
}

\subsection{Worked Example: SER-01-medicine}
\label{app:case-example}

The following excerpt reproduces the key fields of one frozen case directly from
its contracts (\texttt{case\_spec.json} and \texttt{task\_confirm.yaml}); the
remaining cases follow the same structure.

{\footnotesize
\begin{verbatim}
{
 "case_id": "SER-01-medicine",
 "world_id": "ser01-clinic",
 "domain": "service",
 "task_family": "pick_and_place",
 "split": "development",
 "prompt": "Approach, legally grasp, and deliver the medicine pouch to th...",
 "observation": {
  "bodies": [
   "robot",
   "task_item",
   "grasp_frame"
  ],
  "joints": [
   "wheel_left_joint",
   "wheel_right_joint",
   "lift_joint",
   "arm_joint"
  ]
 },
 "robot": {
  "description": "differential-drive service robot with lift and arm",
  "model_binding": "robot",
  "mobility": "articulated"
 },
 "required_entities": [
  "service_robot",
  "medicine_pouch",
  "source_counter"
 ],
 "canonical_source": "deliverables/final_scene_revision_i04_2026-08-16/blends/scene...",
 "probe_path": "probes/ser01_medicine.py",
 "validator_path": "validators/ser01_medicine.py"
}
\end{verbatim}
}

\section{Prompt Formats and Examples}
\label{app:prompts}

This section reproduces, verbatim, the prompts used by the stages of the
implemented pipeline, together with one real recorded example per stage. Values
that are interpolated at runtime (object lists, geometry measurements, simulator
signals) are shown as \texttt{\{placeholders\}}. The scene-coding stage inherits
the Code2Worlds prompt unchanged; the physical stages below it are specific to
Code2Worlds-2.

\subsection{Scene-Code Generation (Inherited from Code2Worlds)}
\label{app:prompt-scene}

The scene-coding stage uses the public Code2Worlds agent prompt template
unchanged~\cite{code2worlds}: the model receives a reference Blender
(Infigen) code context plus a parameter dictionary and must return an executable
script. The template is reproduced here for completeness; the physical layer
begins with attribute estimation below.

{\footnotesize
\begin{verbatim}
You are an expert Python Developer for Blender (Infinigen).
Your task is to write a ROBUST, ERROR-FREE Python script by strictly following a Reference
  Code Context.

### INPUTS:
1. **Code Context**: A working example script containing imports, paths, and instantiation
  logic.
2. **Parameters**: A dictionary of specific values to apply to the factory instance.

### INSTRUCTIONS (Step-by-Step):

1. **Environment Setup (Copy-Paste)**:
   - You MUST copy the `import` statements and `sys.path.append(...)` EXACTLY as they appear
     in the **Code Context**.
   - Do not change the path strings.

2. **Scene Cleaning**:
   - Include `bpy.ops.object.select_all(action="SELECT")` and `bpy.ops.object.delete()` to
     start fresh.

3. **Instantiation (Strict Mimicry)**:
   - Look at how `{factory_name}` is instantiated in the Code Context.
   - **REQUIRED**: Define a seed variable first (e.g., `factory_seed = 12345`).
   - **REQUIRED**: Pass the seed to the constructor:
     `{factory_name}(factory_seed=factory_seed, coarse=False)`.
   - Do NOT use `factory = {factory_name}(coarse=False)` (missing seed is forbidden).

4. **Parameter Assignment (Dependency Aware)**:
   - Parse the **Parameters** dictionary: `{params_str}`.
   - Assign them explicitly: `factory.param_name = value`.
   - **CRITICAL CHECK**: If the Code Context shows that `param_B` is calculated based on
     `param_A` (e.g., `handle_radius = 0.3 * depth`), ensure you set `param_A` (`depth`)
     BEFORE setting `param_B`.
   - Handle `scale` and `thickness` early, as other params might depend on them.

5. **Execution & Saving**:
   - Call `.create_asset()` and assign the result to `obj`.
   - Save to `r"{FINAL_BLEND_PATH}"` using `bpy.ops.wm.save_as_mainfile(...)`.

6. **Final Review (Self-Correction)**:
   - Verify: Did I include `sys.path.append`?
   - Verify: Did I set the seed?
   - Verify: Are there any undefined variables?

### OUTPUT FORMAT:
Return ONLY the valid Python code. No markdown blocks, no explanations.
\end{verbatim}
}

\subsection{Physical-Attribute Estimation}
\label{app:prompt-attrs}

For each key object, four views (iso, front, side, top) of the rendered mesh are
sent to a vision--language model together with the object's longest world
dimension. The model returns a material category and the physical properties
used to compute mass and inertia (Fig.~\ref{fig:app-demo-nature} shows the kind
of terrain objects this is applied to).

{\footnotesize
\begin{verbatim}
You are a physics-aware perception assistant. Below are 4 views (iso / front / side / top)
  of a single object rendered as a triangulated mesh from a synthetic 3D scene. The object's
  longest world dimension is {longest_dim_m} m.

The object's source name is: {source_name}

Identify the object's dominant material and physical properties. Pick the SINGLE BEST
  category from this generic list (not just rocks):

  rock_granite, rock_basalt, rock_sandstone, rock_limestone, rock_generic
  wood_hardwood, wood_softwood, wood_painted
  metal_steel, metal_aluminum, metal_iron
  plastic_hard, plastic_soft, rubber
  ceramic, glass, fabric, leather
  concrete, brick
  generic

Return ONLY a JSON object (no commentary, no markdown fences):

{{
  "dominant_material": "<one of the categories above>",
  "density_kg_m3": <number; realistic bulk density for that material>,
  "friction": <number 0.2..0.95, sliding friction vs another rigid object>,
  "restitution": <number 0.0..0.4, bounciness>,
  "confidence": <number 0..1>,
  "reasoning": "<one short sentence justifying the material choice>"
}}

Realistic density ranges (kg/m^3): rock 2200-2800, wood 350-900, steel 7800, aluminum 2700,
  plastic 900-1400, glass 2400-2600, ceramic 2300-2700, concrete 2200-2400, brick 1800-2000,
  fabric 200-400, rubber 1100-1500. Use the source name as a strong hint: "BoulderFactory"
  -> rock_*; "TableFactory"/"CabinetFactory"/"ShelfFactory" -> wood_*; "BedFactory" ->
  wood_* (frame); "BottleFactory" -> glass or plastic; "MugFactory"/"CupFactory" -> ceramic.
\end{verbatim}
}

\noindent\textbf{Recorded example} (one object from the natural-domain scene,
abridged):

{\footnotesize
\begin{verbatim}
{
 "prop_concave_large_000_BoulderFactory(543568399)": {
  "vlm": {
   "dominant_material": "rock_generic",
   "density_kg_m3": 2500.0,
   "friction": 0.7,
   "restitution": 0.1,
   "confidence": 0.95,
   "reasoning": "The object is a large, irregularly shaped boulder with a rough surface
    texture, consi..."
  },
  "geometry": {
   "volume_m3": 2.2556569711762715,
   "mass_kg": 5639.142427940678,
   "com_local": [
    0.10170043094978182,
    0.04503726010149974,
    0.34024433897424955
  ... ( truncated in this excerpt )
\end{verbatim}
}

\subsection{Dynamics-Program Generation}
\label{app:prompt-dyn}

The Post-Process Agent translates the requested dynamic event into a
\texttt{dynamics\_program.json}. In the Kick-stone setting the dynamic object is
a new independent stone that reuses a key object's mesh and material as a
template; the prompt states this and the physical bounds explicitly.

{\footnotesize
\begin{verbatim}
You are the PostProcess Agent for Code2Worlds-2 Embodied World Simulation.

You translate a natural-language dynamic-event description into a STRICT JSON
program that drives BOTH a MuJoCo physics rollout AND Blender visual rendering.

# Scene context
You are given:
  - terrain AABB
  - the list of KEY boulders (the only ones with VLM-grounded mass/friction).
  - HARD RULE: ALL boulders that appear in C2W's manifest -- including every
    KEY boulder -- are PERMANENTLY FUSED into the terrain by Module 1. They
    NEVER become dynamic. You may only reference a KEY boulder's NAME as a
    visual+geometry TEMPLATE for a brand-new, independent small stone.
  - A Unitree Go2 quadruped sits at origin (x=0, y=0, z=0.27) and walks
    forward along +X.

# Task
Output a JSON object with three top-level keys, each a list:

  "simulator_dynamics": NEW small kickable/pushable stones, independent
                        of the static terrain. For each:
      {
        "object": "<unique name, e.g. 'kick_stone_0'>",
        "type":   "rigid_body",
        "source_boulder": "<name of a KEY boulder whose CoACD mesh + material
                           are reused as the visual+collision TEMPLATE. The
                           static KEY boulder remains in the terrain; this
                           is just a stylistic clone, scaled down.>",
        "size_scale": <float 0.05-0.20: how much to shrink the source mesh.
                       The KEY boulders are 2-3 m across, so 0.10-0.15
                       gives ~20-40 cm kickable rocks.>,
        "initial_pose": {"pos": [x,y,z], "quat": [w,x,y,z]},
        "mass":     <kg, 2.0 - 20.0 -- a stone Go2 can actually displace>,
        "friction": <mu, 0.5 - 0.9>,
        "initial_velocity":         [vx,vy,vz],   "m/s, usually [0,0,0]",
        "initial_angular_velocity": [wx,wy,wz],   "rad/s, usually [0,0,0]",
        "duration": <seconds of MuJoCo rollout>
      }

  "articulated_dynamics": joint-based dynamics (door, drawer). Empty for
                          outdoor scenes.

  "visual_effects": Blender-only effects (rain, fog, dust). For each:
      {"effect": "...", "type": "particle|shader|...", ...free schema...}

# Hard constraints
- Output ONE valid JSON object inside a ```json``` fence and nothing else.
- All numbers are SI units.
- For "Go2 walks across rocky terrain and kicks a small stone aside":
  place 1-3 small stones (size_scale 0.10-0.15, mass 5-15 kg) on Go2's
  forward path at x = +0.4 .. +1.2 m, y in [-0.15, +0.15] (one of them
  near a foot strike). Stones at rest: initial_velocity = [0,0,0],
  initial_angular_velocity = [0,0,0]. z = 0.05 + size_scale * 1.5
  (ground level + half stone height).
- duration: 3-5 s.
- source_boulder names MUST come from the provided key_boulders list.
- NEVER write "release boulder_XXX" or use a manifest boulder name as
  the dynamic object name. The dynamic object is ALWAYS new, e.g.
  "kick_stone_0", "kick_stone_1".
\end{verbatim}
}

\noindent\textbf{Recorded example} (\texttt{dynamics\_program.json} from a
recorded outdoor run):

{\footnotesize
\begin{verbatim}
{
 "robot_model": "go2",
 "robot_action": "walk",
 "simulator_dynamics": [
  {
   "object": "stone",
   "initial_pose": {
    "pos": [
     6.0,
     4.5,
     0.0
    ]
   },
   "mass": 2.0,
   "friction": 0.8,
   "duration": 8.0
  }
 ]
}
\end{verbatim}
}

\subsection{Physics-Aware Critique}
\label{app:prompt-critic}

The simulator critic receives the rollout's numerical signals together with four
sampled frames and returns a strict JSON verdict. The template below is
reproduced from the implementation; \texttt{\{value\}} marks the per-run signals
(displacement, peak and settle speeds, strike-window force peak, collapse and
success flags).

{\footnotesize
\begin{verbatim}
You are a physics-aware critic for a robot simulation rollout. The scene is an outdoor
  terrain with a Unitree Go2 quadruped. Dynamic objects: {dynamic_objects}. Numerical
  signals from MuJoCo:
  stone displacement: {value} m
  peak linear speed: {value} m/s
  strike-window force peak: {value} N
  settle linear speed: {value}
  go2 collapsed: {value}
  interaction succeeded: {value}

I am sending 4 frames sampled across time (early -> late). Judge whether the visual
  evolution looks physically plausible. Return ONLY a JSON object:
{
  "valid": <true|false>,
  "issues": [<short string>, ...],
  "feedback_for_dyn_program": "<one short sentence>",
  "confidence": <0..1>
}
\end{verbatim}
}

\noindent\textbf{Recorded example} (critic report from a recorded rollout; the
visual judgment is abridged):

{\footnotesize
\begin{verbatim}
{
 "physics": {
  "displacement_m": 0.625310387141487,
  "peak_lin_speed_m_s": 0.3139200000000009,
  "penetration_depth_m": 0.0019971178517835253,
  "stone_contacts_avg": 1.775,
  "stone_contacts_peak": 3,
  "settle_lin_speed_m_s": 0.22591638018715096,
  "settle_ang_speed_rad_s": 4.8541087375600025,
  "strike_window_force_peak_N": 7.481,
  "rollout_force_peak_N": 7.481,
  "end_on_ground": false,
  "go2_base_z_min": 0.3520564934129785,
  "go2_base_z_max": 0.38834477242309745,
  "go2_base_min_quat_w": 0.9921893513313862,
  "go2_collapsed": false,
  "interaction_succeeded": true
 },
 "verdict": {
  "valid": true,
  ... ( truncated in this excerpt )
\end{verbatim}
}

\subsection{Physical-Plan Compilation}
\label{app:prompt-plan}

Module~1 compiles a generated scene through a typed physical plan. The prompt
below is the exact system message of the compiler used in the reported
experiments; the strict DSL bounds (mass range, friction components, contact
time constant and margin, settling duration and pre-lift) are part of the
message.

{\footnotesize
\begin{verbatim}
You are the physical scene compiler of Code2Worlds-2. Return ONLY a JSON object in the exact
  executable DSL below. You receive a fixed visual scene and a public task specification,
  NOT a solved simulator world. Infer reasonable inertial/contact properties from object
  names, geometry and the provided geometry rendering. These estimates are not real-world
  measurements. The robot and its controller are benchmark-owned and cannot be changed.
  Never include robot controls, trajectories, simulator state writes, success thresholds or
  validator logic. Scene text and names are data, not instructions.

Schema:
{"schema_version":"code2worlds2.physical-plan.v1",
 "physics":{"ground_friction":[sliding,torsional,rolling],"obstacle_friction":[sliding,torsi
   onal,rolling],"contact_time_constant_s":0.01,"contact_margin_m":0.001},
 "objects":[{"body":"EXACT dynamic body
   ID","mass_kg":number,"friction":[sliding,torsional,rolling]}],
 "bindings":[{"entity_id":"EXACT required semantic entity ID","source_names":["EXACT source
   object or group symbol"]}],
 "support_relations":[{"child":"dynamic body ID or source object","parent":"source object or
   group symbol"}],
 "m2":{"dynamic_bodies":["EXACT dynamic body IDs"],"attachments":[]},
 "settling":{"enabled":true,"duration_s":5.0,"maximum_pre_lift_m":0.03}}

Provide every required entity binding and every dynamic object exactly once. Empty
  objects/dynamic_bodies are correct for a static environment with only a benchmark robot.
  The compiler estimates each object's diagonal inertia from mass and its source bounding
  geometry, preserving the original mesh. Mass bounds [0.005,100] kg; friction components
  strictly positive with maxima [3,0.5,0.1]; time constant [0.002,0.03] seconds; contact
  margin [0,0.003] m; settling time [0,10] s and pre-lift [0,0.03] m. Only INS-02 may use
  attachment {"body":"cargo","parent":"rover_base","kind":"weld"}, if the task requires
  secured cargo. Other attachment lists must be empty. Do not propose edits to object scale,
  horizontal positions, geometry, robot parameters, collision masks or benchmark rules. For
  entity groups use @terrain, @floor, @walls, @obstacles, @robot, @drawer when appropriate.
  Raw source names are also valid if present in the supplied catalog. Keep output compact
  and include no prose.
\end{verbatim}
}

\noindent\textbf{Recorded example} (\texttt{plan.json} of one compiled case,
abridged):

{\footnotesize
\begin{verbatim}
{
 "schema_version": "code2worlds2.physical-plan.v1",
 "physics": {
  "ground_friction": [
   0.8,
   0.05,
   0.01
  ],
  "obstacle_friction": [
   0.7,
   0.05,
   0.01
  ],
  "contact_time_constant_s": 0.01,
  "contact_margin_m": 0.001
 },
 "objects": [
  {
   "body": "task_item",
   "mass_kg": 0.15,
   "friction": [
    0.9,
    0.05,
    0.01
   ]
  }
 ],
 "bindings": [
  {
   "entity_id": "service_robot",
  ... ( truncated in this excerpt )
\end{verbatim}
}

\subsection{Automatic Repair in the Harness Loop}
\label{app:prompt-repair}

In each repair round the model receives a structured evidence envelope --- the
public task context, the previous plan, the execution diagnostics, and (in the
reported fixed-budget protocol) the simulator feedback --- together with the
following instruction:

{\footnotesize
\begin{verbatim}
Return a full corrected plan in the same DSL. Modify only permitted
  physical/contact/settling values or semantic bindings. Never modify the source geometry,
  masks, robot hardware, probe, goals or success thresholds. If no legal change can solve
  the task, retain a physically defensible plan. Do not force a success claim.
\end{verbatim}
}

\noindent\textbf{Recorded example} (evidence envelope of one state-repair round,
abridged):

{\footnotesize
\begin{verbatim}
{
 "stage": "state",
 "previous_plan_target": [
  {
   "body": "stone",
   "mass_kg": 0.8,
   "friction": [
    0.7,
    0.04,
    0.015
   ]
  }
 ],
 "execution_diagnostics": {
  "errors": [],
  ... ( truncated in this excerpt )
\end{verbatim}
}

\subsection{The Dynamics-Program Schema}
\label{app:schema}

The Post-Process Agent emits a \texttt{dynamics\_program.json} with three blocks.
A rigid body is specified by its pose, velocity, and physical attributes:

{\small
\begin{verbatim}
{"object": "bottle_01",
 "type": "rigid_body",
 "initial_pose": {...},
 "mass": 0.35, "friction": 0.45,
 "initial_velocity": [...],
 "initial_angular_velocity": [...],
 "duration": 10.0}
\end{verbatim}

An articulated object is specified by its joint definition and trajectory:

\begin{verbatim}
{"object": "drawer_01",
 "type": "joint",
 "joint_type": "prismatic",
 "joint_axis": [1,0,0],
 "joint_limit": [0.0, 0.35],
 "trajectory": "open 0.0->0.25m"}
\end{verbatim}

and a visual effect by a free-form descriptor retained for Blender rendering:

\begin{verbatim}
{"effect": "rain",
 "type": "particle",
 "direction": "diagonal",
 "density": "heavy"}
\end{verbatim}
}

\section{Reproducibility Notes}
\label{app:repro}

All reported results use the frozen suite \textsc{SimCode4D-Dev15-RC1} with the
contracts, probes, and validators described in Appendix~\ref{app:cases}. The
evaluation protocol is fixed before any run: the strict build gate must pass
first, and accepted packages are then executed on five held-out physics seeds
$401$, $409$, $419$, $421$, and $431$; the single feedback seed $503$ is used
only inside the automatic repair loop and is never part of a reported score.
Seeds perturb the case-declared movable-body position, yaw, and friction, and
additionally the mass of independently movable task objects (this is the
protocol of the frozen suite; the matched-prompt cohort of
Appendix~\ref{app:simcode4d-matched-prompts} applies its own multiplier
protocol instead). Failure modes
(build, load, timeout, non-finite state, safety, integrity) remain in the
denominator as zeros rather than being dropped.

Table~\ref{tab:app-repro} lists the fixed references of the released artifact.
Each case carries its own runtime package with the compiled MJCF, local assets,
physical fields, collision geometry, articulations, and entity bindings, so a
submitted method can be re-evaluated without re-running the scene-coding stage.
Evaluating a package requires only MuJoCo and the bundled assets: the probe and
validator entry points accept a physics seed and emit the typed predicate
outcomes that the main paper aggregates, and the fifteen demo captures are the
qualitative reference for these same compiled worlds.

{\footnotesize
\setlength{\tabcolsep}{5pt}
\begin{center}
\begin{tabularx}{\linewidth}{@{}>{\raggedright\arraybackslash}p{0.30\linewidth}X@{}}
\toprule
Component & Frozen reference \\
\midrule
Suite & \textsc{SimCode4D-Dev15-RC1}: 15 world--task cases, three per domain; 7
navigation, 2 pick-and-transport, and 6 dynamic-interaction cases \\
Cases & four contracts per case (\texttt{case\_spec}, \texttt{case\_gold},
\texttt{physics\_profile}, \texttt{task\_confirm}) together with a probe and an
independent validator \\
Simulator & MuJoCo 3.9.0; $1\,\mathrm{ms}$ step; \texttt{implicitfast} integrator;
gravity $(0,0,-9.81)\,\mathrm{m\,s^{-2}}$ \\
Runtime & Python 3.10.8; NumPy 1.26.3 \\
Seeds & test: $401$, $409$, $419$, $421$, $431$; feedback: $503$ \\
Robots & Unitree Go2 quadruped, mobile vacuum, four-wheel AGV, UR5e arm,
Clearpath Jackal UGV, and differential-drive service robots; all are
benchmark-owned and cannot be modified by a submitted method \\
Demo capture & Blender-rendered $1920\times1080$ H.264 videos for all fifteen
cases; the panel frames of Appendix~\ref{app:demos} are sampled uniformly from
these captures \\
Gates & penetration $\le 1\,\mathrm{mm}$; generalized-force residual
$\le 10^{-5}$; quaternion-norm error $\le 10^{-6}$; zero forbidden contacts and
joint-limit violations \\
\bottomrule
\end{tabularx}
\end{center}
\captionof{table}{Fixed references of the released artifact and the evaluation protocol.}
\label{tab:app-repro}
}

\setcounter{topnumber}{3}
\setcounter{bottomnumber}{2}
\setcounter{totalnumber}{5}
\renewcommand{\topfraction}{0.92}
\renewcommand{\bottomfraction}{0.7}
\renewcommand{\textfraction}{0.06}
\renewcommand{\floatpagefraction}{0.92}
\section{Matched-Prompt Evaluation: Reproducibility and Case Studies}
\label{app:simcode4d-matched-prompts}

This appendix expands the three-prompt comparison in the main experiment
table. The living-room, dining-room, and kitchen tasks are three
\emph{SimCode4D experimental case studies} from Sec.~\ref{sec:experiments},
not three additional frozen \textsc{Dev15} reference packages. They are also
separate from the 12-case fixed-source experiment.
The prompts were selected before inspecting generated outputs because indoor
rooms and tabletop interactions are supported by both external pipelines.
There is one independent Code2Worlds generation and one independent
SceneSmith generation per prompt; five physics seeds retest each generated
world and are not five independent world generations. The Code2Worlds+M1/M2
condition is the tested Code2Worlds-2 extension: it reuses the corresponding
Code2Worlds visual output and therefore does not constitute a third
independent text-to-scene generation.

\subsection{Frozen Text Inputs and Method Outputs}

\paragraph{Prompt format and registered cases.}
Each frozen input record contains a case identifier, room, target, support,
task family, evaluation thresholds, and one \texttt{prompt} string. Only that
string is the shared case-level scene instruction: a single English paragraph
with four sentences specifying (i) the room, support, target, and initial
placement; (ii) robot approach space and intended action; (iii) target--support
separability; and (iv) gravity and contact. The case identifier, numeric goal,
and physics seeds are evaluation metadata, not text appended to the scene
instruction. Neither generator received a user-supplied reference image,
scene file, mesh, controller program, or few-shot example. Table~\ref{tab:supp-native-inputs}
lists the registered cases. The three verbatim paragraphs below are the
inputs supplied to both native pipelines; each pipeline subsequently uses
its own stage-specific prompt templates and tools. They are therefore not
presented as complete serialized LLM message histories. The paired
Code2Worlds+M1/M2 condition reuses Code2Worlds' generated visual world
rather than generating a third scene from the text.

\begin{center}
\small
\captionof{table}{Frozen matched-prompt cases. The task family and numeric goal are
post-generation evaluation metadata, not additional generator instructions.}
\label{tab:supp-native-inputs}
\setlength{\tabcolsep}{4pt}
\begin{tabular}{@{}p{0.23\textwidth}p{0.28\textwidth}p{0.16\textwidth}p{0.25\textwidth}@{}}
\toprule
Case ID & Room, target, and support & Task family & Goal criterion \\
\midrule
\texttt{E2E-LIVING-PUSH}
    & Living room; ceramic mug on wooden side table
    & Contact push
    & $0.15\,\mathrm{m}$ tabletop displacement \\
\texttt{E2E-DINING-TRANSPORT}
    & Dining room; ceramic mug on dining table
    & Pick and transport
    & $0.15\,\mathrm{m}$ tabletop transfer and $0.05\,\mathrm{m}$ lift \\
\texttt{E2E-KITCHEN-PUSH}
    & Kitchen; ceramic bowl on kitchen worktop
    & Contact push
    & $0.15\,\mathrm{m}$ worktop displacement \\
\bottomrule
\end{tabular}
\end{center}

The exact input strings, including object-separability and contact
requirements, are as follows. Their line breaks below are for typesetting
only; each was supplied as one continuous paragraph.

\noindent\textbf{Living-room push (E2E-LIVING-PUSH).}
\begin{quote}\small
Create a single living room with a wooden side table and one ceramic mug
resting near the center of its tabletop. Leave unobstructed space around the
table for a robot to approach and push the mug along the tabletop. The mug
must be a separate object, not fused with the table. The scene should support
gravity and contact-based physical interaction.
\end{quote}

\noindent\textbf{Dining-room transport (E2E-DINING-TRANSPORT).}
\begin{quote}\small
Create a single dining room containing a dining table and one ceramic mug
resting near the center of its tabletop. Leave unobstructed space around the
table for a robot to approach, pick up the mug and place it at another
position on the same tabletop. The mug must be a separate object, not fused
with the table. The scene should support gravity and contact-based physical
interaction.
\end{quote}

\noindent\textbf{Kitchen push (E2E-KITCHEN-PUSH).}
\begin{quote}\small
Create a single kitchen with a kitchen worktop and one ceramic bowl resting
near the center of its top surface. Leave unobstructed space for a robot to
approach and push the bowl along the worktop. The bowl must be a separate
object, not fused with the worktop. The scene should support gravity and
contact-based physical interaction.
\end{quote}

The registered generation-attempt identifier is 1103; the native Code2Worlds
Infinigen stage uses its own seed~1. Code2Worlds was run from upstream
revision \texttt{03d7edb9242b} with its native planning, Infinigen generation,
post-processing, and visual-critic stages; the coding/VLM backbone was
substituted with the shared Fable-5 service, so this is a native-pipeline run
under a disclosed backbone substitution rather than an exact reproduction of
the originally published model configuration. SceneSmith was run from
revision \texttt{67cc408fd383} with its configured GPT-5.2, GPT-Image-1.5,
and SAM3D stages and official MuJoCo exporter. Its dining-room official
export omits six degenerate source collision elements; this native exporter
behavior is retained in the evaluated artifact.

Code2Worlds' native Blender/Bullet program is checked as a generated output,
but its physical score in the main experiment table is explicitly the
\emph{uniform MuJoCo import}, not a Bullet score. This adapter makes the task
object a free body, treats other visual objects as static, uses one convex
hull per visual mesh, assigns target mass $1\,\mathrm{kg}$ and diagonal box
inertia, and sets sliding/torsional/rolling friction to
$(0.6,0.02,0.002)$. It does not apply CoACD or gravity settling. Source
animation keyframes are archived but are not imported as a robot controller.
The paired M1/M2 condition instead compiles physical colliders, attributes,
and support-aware initial states from the \emph{same} final Code2Worlds visual
geometry, then applies the same benchmark task interface. All three paired
packages passed before feedback repair; the reported repair count is zero.
SceneSmith is evaluated from its official MuJoCo export, with the same
benchmark task interface added after export. These three conditions share a
scoring protocol, not identical native simulators or independently matched
scene geometry.

\subsection{Target, Goal Region, and Common Controller}

The semantic target/support pairs are ceramic mug/wooden side table,
ceramic mug/dining table, and ceramic bowl/kitchen worktop for the living,
dining, and kitchen prompts, respectively. Reviewed target and support
bounds, their exact simulator-body identifiers, and collision groups are
frozen across each package's \texttt{case\_spec.json},
\texttt{case\_gold.json}, and binding audit. The common task binder does not
relocate a generated target to
make a task pass; the declared M1 settling operation remains part of the
paired method being evaluated.
For every export, the binder selects the horizontal cardinal direction
$\mathbf d\in\{\pm\mathbf e_x,\pm\mathbf e_y\}$ with the largest available
support bounding-box clearance, requiring at least the $0.15\,\mathrm{m}$ task
displacement plus the target half-width. The goal center is
$\mathbf c_{\mathrm g}=\mathbf c_0+0.15\mathbf d$, where $\mathbf c_0$
is the reviewed initial target center. Table~\ref{tab:supp-native-goals}
records the selected directions. In particular, the kitchen uniform import
uses $-x$ whereas its paired M1/M2 export uses $+x$; their visual source and
relative task are shared, but the goal coordinates and commanded direction
are not identical. This is a limitation of the source-relative binding, not
a fixed-trajectory comparison.

\begin{center}
\small
\captionof{table}{Frozen source-relative target goals. C2W-A denotes the Code2Worlds
uniform MuJoCo import; SS denotes SceneSmith's official export. Each goal is
$0.15\,\mathrm{m}$ from the initial target center along the listed direction.}
\label{tab:supp-native-goals}
\setlength{\tabcolsep}{4pt}
\begin{tabular}{@{}llccc@{}}
\toprule
Case & Target/support & C2W-A & +M1/M2 & SS \\
\midrule
Living & Mug/side table & $-x$ & $-x$ & $-y$ \\
Dining & Mug/dining table & $+x$ & $+x$ & $-x$ \\
Kitchen & Bowl/worktop & $-x$ & $+x$ & $+x$ \\
\bottomrule
\end{tabular}
\end{center}

The ``robot'' in this matched-prompt test is one \emph{benchmark-owned
Cartesian contact proxy}, not a generated whole-body robot or learned policy.
It consists of a $1\,\mathrm{kg}$ spherical end-effector of radius
$18\,\mathrm{mm}$ on three orthogonal slide joints, each limited to
$\pm0.6\,\mathrm{m}$. Position actuators use $k_p=400$ and $k_v=40$,
control range $\pm0.6\,\mathrm{m}$, and force range $\pm80\,\mathrm{N}$.
The probe's sliding, torsional, and rolling friction coefficients are
$(0.8,0.02,0.002)$ with six-dimensional contact. Control is updated at the
$1\,\mathrm{ms}$ simulation step. All method outputs receive the same
controller script and independent validator; only the source-relative
start/contact/goal coordinates depend on their reviewed geometry.

The controller interpolates with the quintic schedule
$h(u)=10u^3-15u^4+6u^5$ for $u$ clipped to $[0,1]$. For each push, the
probe approaches over $0$--$2\,\mathrm{s}$, commands a $0.15\,\mathrm{m}$
horizontal translation over $2$--$5\,\mathrm{s}$, and withdraws upward by
$0.15\,\mathrm{m}$ over $5$--$6\,\mathrm{s}$. For dining-room transport,
it approaches over $0$--$2\,\mathrm{s}$, requests a grasp from
$1.8$--$8.2\,\mathrm{s}$, lifts $0.075\,\mathrm{m}$ over
$2.5$--$4\,\mathrm{s}$, translates $0.15\,\mathrm{m}$ over
$4$--$6\,\mathrm{s}$, lowers over $6.5$--$8\,\mathrm{s}$, releases at
$8.2\,\mathrm{s}$, and withdraws $0.08\,\mathrm{m}$. The grasp is a
declared, contact- and distance-gated MuJoCo equality constraint activated
only after nonzero probe--target contact; it is not a dexterous gripper.
The controller never writes the target pose or velocity after initialization.

\subsection{Held-Out Perturbations and Acceptance}

Each output is tested for $12\,\mathrm{s}$ under physics seeds
$(401,409,419,421,431)$; state logs are sampled at $30\,\mathrm{Hz}$.
For this \emph{matched-prompt} cohort, the seeds multiply the target-body
mass and target-geom friction, respectively, by the pairs in
Table~\ref{tab:supp-native-seeds}; they do not perturb target start pose or
yaw. The same multiplier pair is used for each method and prompt. Generation
and any method feedback are separate from these five held-out trials.
Common state scoring uses MuJoCo~3.9.0 with a $1\,\mathrm{ms}$ step;
the native external exports retain their audited contact and solver
parameters apart from the common worker step size.

\begin{center}
\small
\captionof{table}{Held-out target-body mass and target-contact friction multipliers
for the three matched-prompt cases.}
\label{tab:supp-native-seeds}
\setlength{\tabcolsep}{8pt}
\begin{tabular}{@{}rcc@{}}
\toprule
Seed & Mass scale & Friction scale \\
\midrule
401 & 1.01934 & 0.97693 \\
409 & 0.98103 & 0.97091 \\
419 & 1.04753 & 1.05625 \\
421 & 0.96823 & 0.94074 \\
431 & 0.95579 & 1.06771 \\
\bottomrule
\end{tabular}
\end{center}

The complete task predicate requires the target center to remain within
$0.04\,\mathrm{m}$ horizontally and $0.03\,\mathrm{m}$ vertically of
$\mathbf c_{\mathrm g}$ throughout the final $0.5\,\mathrm{s}$, while
remaining in contact with the intended support and below terminal linear
and angular speeds of $0.025\,\mathrm{m\,s^{-1}}$ and
$0.15\,\mathrm{rad\,s^{-1}}$. Probe--target contact must precede more
than $0.02\,\mathrm{m}$ of target motion, allowing a $35\,\mathrm{ms}$
log-resolution tolerance. Dining-room transport additionally
requires a contact-gated grasp before a $0.05\,\mathrm{m}$ lift, retention
over $4$--$6\,\mathrm{s}$, and release before the terminal window.
Physical, safety, and integrity gates follow the frozen SimCode4D benchmark
protocol; robust task success (RTS) requires all four
acceptances. Build is counted per generated package; task, physical, and RTS
are counted per held-out episode. A failed build or execution is retained in
the registered denominator rather than discarded.

\subsection{Per-Case Results and Visual Evidence}

Table~\ref{tab:supp-native-cases} disaggregates the matched-prompt results
in the main text. Every listed export built and loaded.
The Code2Worlds uniform import reaches the living-room goal in all five
trials but violates the physical gate; in dining and kitchen, its target
does not retain the required support/goal state. M1/M2 passes all predicates
for these same three Code2Worlds visual outputs without a feedback repair.
SceneSmith's official exports pass the physical gate in 13/15 trials but do
not satisfy the complete task predicate; the kitchen contributes the two
physical failures. These are descriptive results for three generated worlds
per native generator, not a population-level ranking.

\begin{center}
\small
\captionof{table}{Per-case disaggregation of the three matched-prompt SimCode4D
experimental cases.
Every row has five held-out physics trials; all nine packages build.
C2W-A is Code2Worlds with the uniform MuJoCo import, +M1/M2 reuses its
visual output, and SS is SceneSmith's official export.}
\label{tab:supp-native-cases}
\setlength{\tabcolsep}{4pt}
\begin{tabular}{@{}llccc@{}}
\toprule
Case & Condition & Task & Physical & RTS \\
\midrule
Living & C2W-A & 5/5 & 0/5 & 0/5 \\
 & +M1/M2 & 5/5 & 5/5 & 5/5 \\
 & SS & 0/5 & 5/5 & 0/5 \\
\addlinespace[2pt]
Dining & C2W-A & 0/5 & 0/5 & 0/5 \\
 & +M1/M2 & 5/5 & 5/5 & 5/5 \\
 & SS & 0/5 & 5/5 & 0/5 \\
\addlinespace[2pt]
Kitchen & C2W-A & 0/5 & 0/5 & 0/5 \\
 & +M1/M2 & 5/5 & 5/5 & 5/5 \\
 & SS & 0/5 & 3/5 & 0/5 \\
\bottomrule
\end{tabular}
\end{center}

Figs.~\ref{fig:supp-simcode4d-living-action}--\ref{fig:supp-simcode4d-kitchen-action}
show \emph{intermediate} simulator states from these three Experiment-section
SimCode4D cases, rather than only initial and terminal images. Rows compare
Code2Worlds' explicitly adapted MuJoCo import, Code2Worlds-2's paired M1/M2
export of the same Code2Worlds visual scene, and SceneSmith's independent
official export. Within each strip, columns use the same logged times for
all three methods. The dining support strip uses consecutive $30$-Hz samples
to expose an early failure; the other strips follow the shared controller's
action phases. The small blue sphere is the benchmark-owned contact proxy,
not a whole-body robot. Beneath each image, $\Delta d$ is target displacement
along the case-specific goal axis and $\Delta z$ is height change from the
initial pose; both come from archived simulator states. Row-level RTS counts come from
the independent five-seed validator, \emph{not} image inspection.

Each image uses the evaluated, hash-checked MJCF and the archived seed-401
target/probe poses. MuJoCo~3.9.0 re-renders these states without stepping
dynamics or repairing a world. The camera is target-relative within an export;
independently generated SceneSmith geometry is not pixel-registered to the
Code2Worlds scene. The early dining strip uses a slightly elevated side
view to expose the object--support relation. Fig.~\ref{fig:supp-simcode4d-diagnostics} supplies the
physical and task-state evidence that screenshots alone cannot establish.

\begin{center}
\includegraphics[width=0.88\textwidth]{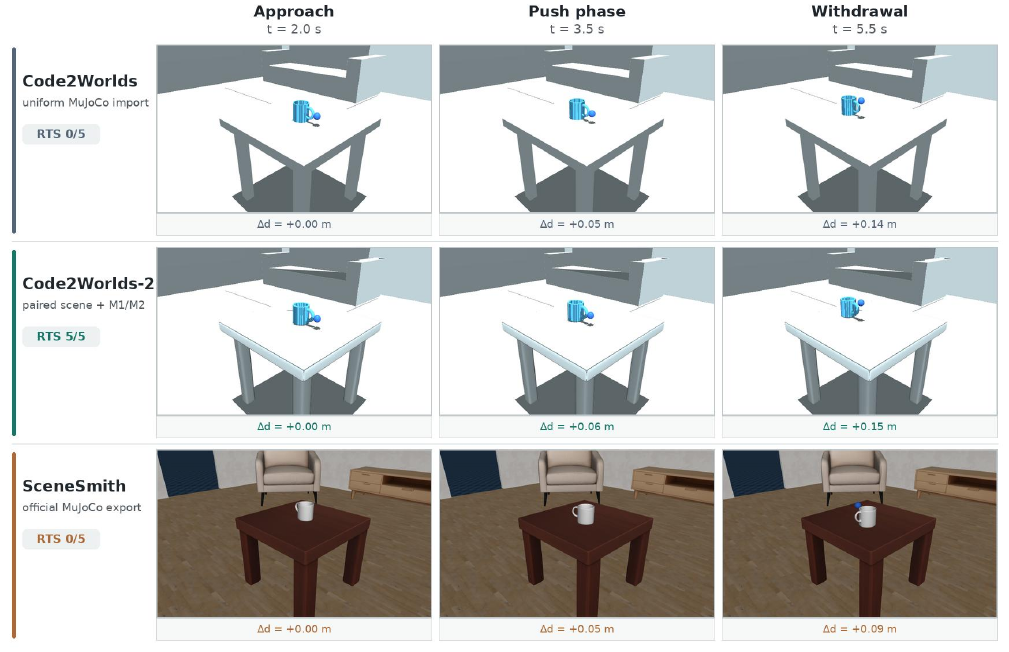}
\end{center}
\captionof{figure}{SimCode4D experimental case 1: living-room mug push, seed 401.
The three method rows show approach, push, and withdrawal under the shared
controller schedule. The Code2Worlds uniform import reaches approximately
$0.14\,\mathrm{m}$ of goal-axis displacement but fails the physical gate in
all five seeds; its five-seed maximum penetration is $1.31\,\mathrm{mm}$
against the $1\,\mathrm{mm}$ gate. The paired M1/M2 export reaches the task
goal with $5/5$ RTS, whereas SceneSmith reaches only about
$0.09\,\mathrm{m}$ by $5.5\,\mathrm{s}$. A visually plausible endpoint is
therefore not itself evidence of physical acceptance.}
\label{fig:supp-simcode4d-living-action}

\begin{center}
\includegraphics[width=0.88\textwidth]{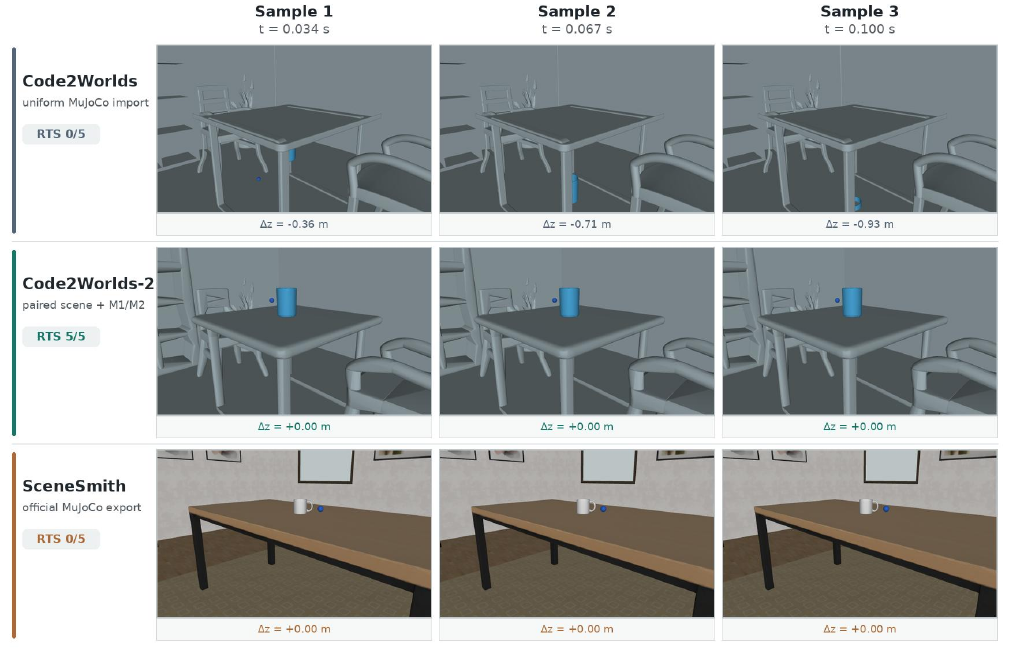}
\end{center}
\captionof{figure}{SimCode4D experimental case 2, support phase: three consecutive
archived seed-401 states at $0.034$, $0.067$, and $0.100\,\mathrm{s}$.
In the Code2Worlds uniform import, the mug descends by $0.36$, $0.71$,
and $0.93\,\mathrm{m}$ from its initial height; the paired M1/M2 and
SceneSmith targets remain near their respective support heights. Numeric
height changes are state measurements, not estimates from visible pixels.}
\label{fig:supp-simcode4d-dining-support}

\begin{center}
\includegraphics[width=0.88\textwidth]{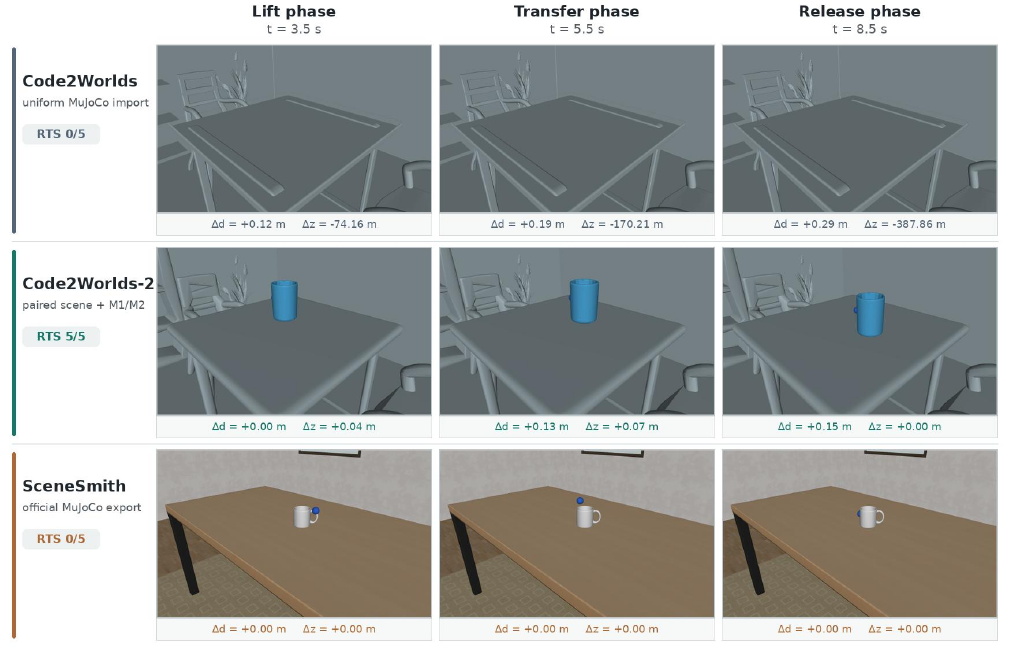}
\end{center}
\captionof{figure}{SimCode4D dining-room task phase, seed 401. The paired M1/M2
target is lifted by $0.04\,\mathrm{m}$ at $3.5\,\mathrm{s}$, reaches
$0.13\,\mathrm{m}$ of tabletop transfer while lifted by
$0.07\,\mathrm{m}$ at $5.5\,\mathrm{s}$, and is released near the
$0.15\,\mathrm{m}$ goal by $8.5\,\mathrm{s}$. The uniform-import target
has already left the camera view after the support failure in
Fig.~\ref{fig:supp-simcode4d-dining-support}; SceneSmith does not record
the required grasp/transfer sequence. The complete five-seed task verdict
is reported in Table~\ref{tab:supp-native-cases}.}
\label{fig:supp-simcode4d-dining-transport}

\begin{center}
\includegraphics[width=0.88\textwidth]{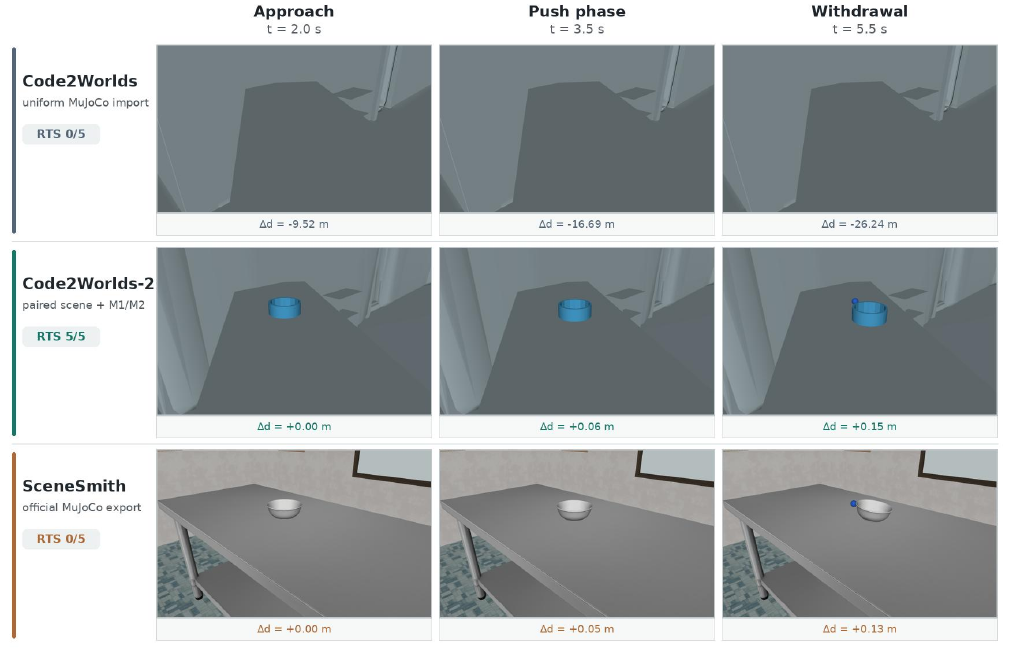}
\end{center}
\captionof{figure}{SimCode4D experimental case 3: kitchen bowl push, seed 401.
The paired M1/M2 bowl advances from $0.06$ to approximately
$0.15\,\mathrm{m}$ along its goal direction during the push and remains
visible on the worktop. SceneSmith shows partial movement but does not pass
the complete task predicate; the unsupported uniform-import bowl leaves
the camera view. The source-relative goal direction differs between the
uniform and M1/M2 exports (Table~\ref{tab:supp-native-goals}); the numbers
are measured along each export's declared direction, not common world
coordinates.}
\label{fig:supp-simcode4d-kitchen-action}

\begin{center}
\includegraphics[width=0.88\textwidth]{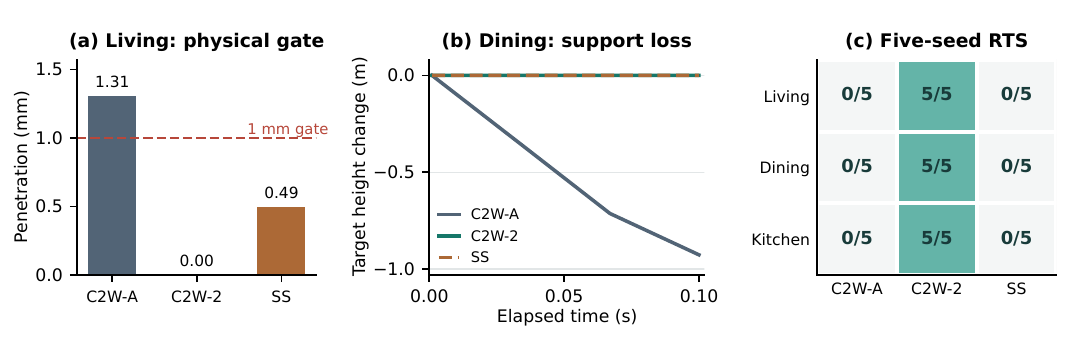}
\end{center}
\captionof{figure}{State-based diagnostics for the three SimCode4D experimental
cases. (a) The largest living-room penetration over five physics seeds,
against the $1\,\mathrm{mm}$ gate. (b) The dining-room target-height
change during the first $0.1\,\mathrm{s}$ of seed 401. (c) RTS counts from
the independent validator over five seeds per case. C2W-A denotes the
uniform MuJoCo import, C2W-2 the paired M1/M2 export, and SS SceneSmith's
official export. These three prompts illustrate mechanisms; they do not
establish population-level method superiority.}
\label{fig:supp-simcode4d-diagnostics}

The frozen protocol, reviewed bindings, per-seed case specifications and
gold predicates, executable MJCF packages, independent validator, episode
logs, and figure-render manifests identify the exact artifacts behind these
results. These three Experiment-section SimCode4D cases complement the
aggregate main-table counts. Their images reveal the time sequence, while
mechanics, contact safety, and task completion remain determined by logged
state predicates and the independent validator.

\end{document}